\pdfoutput=1
\documentclass[10pt,logo,copyright]{nvidiatechreport}
\usepackage[authoryear,sort&compress,square]{natbib}
\usepackage{minted}
\usepackage{bbm}
\usepackage{caption}

\newcommand{\nickname}{OpenVoxel\xspace}

\usepackage{tikzducks}
\usepackage{graphicx}

\usepackage{multirow}

\usepackage{wrapfig}

\usepackage[nameinlink]{cleveref}
\crefname{equation}{Eq.}{Eqs.}
\crefname{figure}{Fig.}{Figs.}
\crefname{section}{Sec.}{Sec.}
\crefname{appendix}{App.}{App.}
\crefname{table}{Tab.}{Tabs.}
\crefname{algorithm}{Alg.}{Alg.}
\crefname{thm}{Thm}{Thm}
\Crefname{thm}{Thm}{Thm}
\crefname{prop}{Prop}{Prop}
\crefname{line}{Line}{Lines}
\usepackage{longtable}

\usepackage{listings}
\usepackage{xcolor}

\lstdefinestyle{promptcode}{
  basicstyle=\ttfamily\small,
  frame=single,
  numbers=left,
  numberstyle=\tiny,
  stepnumber=1,
  numbersep=8pt,
  breaklines=true,
  breakatwhitespace=false,
  columns=fullflexible,
  keepspaces=true,
  showstringspaces=false,
  tabsize=2
}

\lstnewenvironment{promptlisting}[1][]
{\lstset{style=promptcode,#1}}
{}

\let\cite\citep  

\newcommand{\crefnames}[3]{%
  \@for\next:=#1\do{%
    \expandafter\crefname\expandafter{\next}{#2}{#3}%
  }%
}

\title{\nickname: Training-Free Grouping and Captioning Voxels for Open-Vocabulary 3D Scene Understanding}

\author{
Sheng-Yu Huang$^{1,2,\dagger}$, Jaesung Choe$^1$, Frank Wang$^{1,2}$, Cheng Sun$^1$\\
$^1$NVIDIA, $^2$National Taiwan University\\
}

\begin{abstract}
We propose \textbf{\textit{\nickname}}, a training-free algorithm for grouping and captioning sparse voxels for the open-vocabulary 3D scene understanding tasks. Given the sparse voxel rasterization (SVR) model obtained from multi-view images of a 3D scene, our \nickname is able to produce meaningful groups that describe different objects in the scene. Also, by leveraging 
powerful Vision Language Models (VLMs) and Multi-modal Large Language Models (MLLMs), our \nickname successfully build an informative scene map by captioning each group, enabling further 3D scene understanding tasks such as open-vocabulary segmentation (OVS) or referring expression segmentation (RES). Unlike previous methods, our method is training-free and does not introduce embeddings from a CLIP/BERT text encoder. Instead, we directly proceed with text-to-text search using MLLMs. Through extensive experiments, our method demonstrates superior performance compared to recent studies, particularly in complex referring expression segmentation (RES) tasks. The code will be open.

\end{abstract}

\begin{document}
\maketitle
\abscontent

\section{Introduction}
\label{sec:introduction}
With the recent advancement of the neural rendering algorithm~\cite{mildenhall2021nerf, barron2021mip, yu2021plenoctrees, muller2022instant, sun2022direct, fridovich2022plenoxels, martin2021nerf, reiser2021kilonerf, chen2022tensorf, chou2024gsnerf, liu2022neural, vora2021nesf, johari2022geonerf, suhail2022generalizable, kundu2022panoptic}, several primitives have been introduced to represent the 3D scenes. 3D Gaussian Splatting (3DGS)~\cite{kerbl202333dgs} and several extensions~\cite{chen2024surveygs, gao2023relightable, yu2024mipsplat, xu2025depthsplat} have recently emerged as a groundbreaking technique for novel view synthesis, offering an exceptional balance of visual fidelity and real-time performance. 
Concurrently, Sparse Voxel Rasterization (SVR)~\cite{svr, li2025geosvr} enables a sparse voxel representation to perform the novel-view synthesis task.

Recent studies~\cite{3dovs, lerf, langsplat, refersplat, lee2025mosaic3d, ye2023gaussiangrouping, jun2025drsplat, tian2025ccllgs, 3dgsgrounding, wu2024opengaussian, huang2025openinsgaussian, piekenbrinck2025opensplat3d, he2025pointseg, zhu2025objectgs, jang2025identityawaregs, peng20243dvlgs, ying2024omniseg3d, ji2025fastlgs, marrie2025ludvig} leverage the series of novel 3D primitives for open-vocabulary 3D understanding tasks, such as semantic segmentation, 3D open-vocabulary segmentation (OVS).
\begin{figure}[!t]
    \centering
    \includegraphics[width=0.7\linewidth]{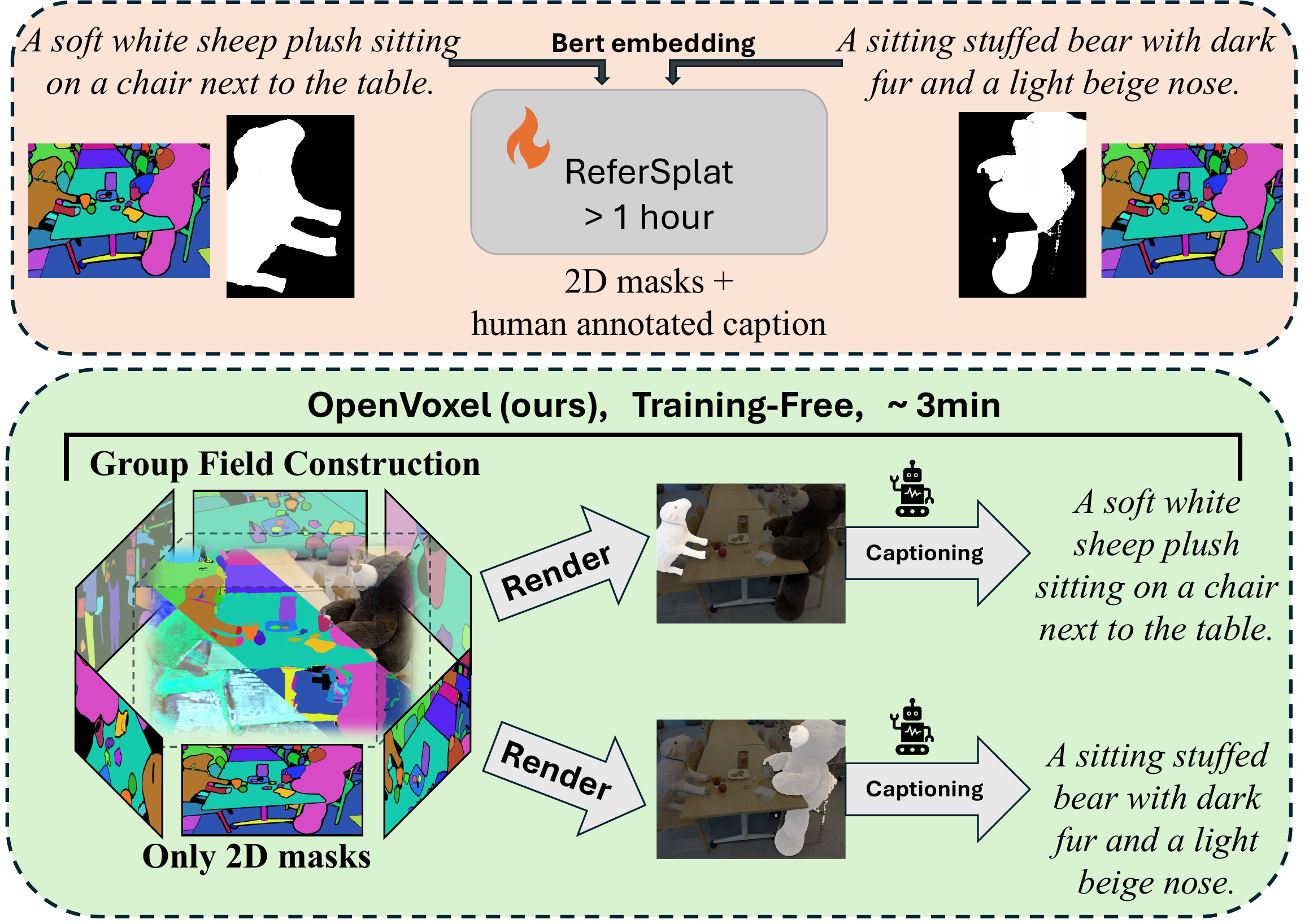}
    \caption{\textbf{Comparison of lifting language into 3D representations between  ReferSplat~\cite{refersplat} and \nickname.} Note that ReferSplat additionally requires manual text annotations for training 3DGS to equipped with Bert embeddings, while ours is totally training-free and do not need any additional annotations.}
    \label{fig:teaser}
\end{figure}

LangSplat~\cite{langsplat} is one of the pioneering papers that introduces 3D Gaussians for this purpose by optimizing 3DGS first and adopts Gaussian features that are trained to store language attributes derived from the 2D foundation model, CLIP~\cite{clip}. The successive studies~\cite{zhu2025objectgs, jang2025identityawaregs, piekenbrinck2025opensplat3d, ying2024omniseg3d, ye2023gaussiangrouping} try to improve the trainable Gaussian features by either introducing contrastive learning between different objects in the scene, or training an object-aware 3DGS model from scratch. However, these methods are limited by learning short words or tags, which hinders advanced 3D scene understanding with complex sentence queries such as the referring segmentation task, as mentioned in~\cite{refersplat}. More recently, ReferSplat~\cite{refersplat} proposes to cope with segmentation from complex queries, and focus on the 3D referring segmentation task (RES) by providing human-annotated observable objects' masks and corresponding detailed natural language sentence descriptions for training. However, all the aforementioned approaches tend to leverage learned embeddings as language fields from visual-language aligned embeddings such as CLIP or DINO, limiting the capability of dealing with arbitrary textual queries~\cite{kamath2024hard}. Also, obtaining detailed human annotation for training requires labor-intensive pre-processing for each scene, and training additional language fields for 3D representation is also time-consuming (more than one hour for each scene).




To address these issues, we propose \textbf{\textit{\nickname}}, a novel training-free framework that bypasses the speed and accuracy limitations of learning text embedding space. Instead of training 3D features to align with a fixed embedding manifold, our \nickname directly populates the 3D scene with rich, human-readable text captions and proceeds with text-to-text retrieval for open-vocabulary 3D scene understanding, including open-vocabulary segmentation (OVS) and referring expression segmentation (RES). Given a Sparse Voxel Rasterization (SVR) model optimized from multi-view images of a 3D scene, we introduce a training-free algorithm of \textit{Training-Free Sparse Voxel Grouping} to cluster all voxels into object-level groups by lifting and matching per-frame 2D segmentation maps obtained from SAM2~\cite{ravi2024sam2}. This aggregation is crucial for moving from a per-voxel representation to a semantically coherent, object-level understanding and ensuring view-consistency. To further bridge voxels with textual description, we also introduce the strategy of \textit{Canonical Scene Map Construction} to locate language attributes on top of this sparse voxel representation via Vision Language Models (VLMs) and Multi-modal Large Language Models (MLLMs), obtaining a full description of the entire scene. Finally, to answer an open-vocabulary query (e.g., a simple word or a complex sentence), we utilize MLLMs to perform \textit{Referring Query Inference} via direct text-to-text retrieval. The MLLM compares the user's query against the constructed scene map, enabling a flexible and powerful retrieval process that is not constrained by embedding proximity.   
In summary, the contributions of our methods are listed below:
\begin{itemize}
    \item \emph{A training-free framework for open-vocabulary 3D scene understanding}~:~We introduce \textbf{\textit{\nickname}}, a novel pipeline that removes the need for learning 3D language embeddings and instead performs text-to-text reasoning using LLMs on top of the sparse voxel representation.
    \item \emph{Group (object)-level captioning and scene map construction}~:~We propose an unsupervised method that clusters sparse voxels into instance-level 3D masks and generates rich, descriptive captions for each instance using an MLLM, forming an informative scene map.
    \item \emph{Superior performance in both complex and simple language queries for 3D scene understanding task}~:~By combining instance captions with LLM-based reasoning, \nickname outperforms previous studies on both open-vocabulary semantic segmentation (OVS) and referring expression segmentation (RES) tasks.
\end{itemize}

\section{Related Works}
\label{sec:related_works}

\noindent\textbf{3D scene representation with different primitives.}
Neural radiance fields (NeRF)~\cite{mildenhall2021nerf, barron2021mip, yu2021plenoctrees, muller2022instant, sun2022direct, fridovich2022plenoxels, martin2021nerf, reiser2021kilonerf, chen2022tensorf, chou2024gsnerf, liu2022neural, vora2021nesf, johari2022geonerf, suhail2022generalizable, kundu2022panoptic} have established a powerful paradigm for representing scenes as continuous volumetric fields, enabling high-quality novel view synthesis but suffering from slow training and rendering. To address these limitations, 3D Gaussian Splatting (3DGS)~\cite{kerbl202333dgs} and following extensions~\cite{chen2024surveygs, gao2023relightable, yu2024mipsplat, xu2025depthsplat} represent scenes using anisotropic Gaussians that can be rendered in real time, significantly improving efficiency while preserving fidelity. More recent work on Sparse Voxel Rasterization (SVR)~\cite{svr} introduces a discrete, sparse voxel representation that supports efficient rasterization-based rendering. These representations (i.e., NeRF, 3DGS, and SVR) collectively form the foundation for modern neural scene reconstruction and are increasingly adopted for downstream 3D understanding tasks.

\smallskip
\noindent\textbf{Language-aligned 3D scene representation.}
A growing line of research~\cite{3dovs, lerf, langsplat, refersplat, lee2025mosaic3d, ye2023gaussiangrouping, jun2025drsplat, tian2025ccllgs, 3dgsgrounding, wu2024opengaussian, huang2025openinsgaussian, piekenbrinck2025opensplat3d, he2025pointseg, zhu2025objectgs, jang2025identityawaregs, peng20243dvlgs, ying2024omniseg3d, ji2025fastlgs} aims to equip 3D scene representations with open-vocabulary or language-aware capabilities so that further downstream tasks~\cite{huang20253dgic, mirzaei2023spin, mirzaei2023referenceinpaint, weder2023removingnerf, yin2023ornerf, chen2024mvip, lin2024maldnerf, wang2024gscream, weber2024nerfiller, wang2024innerf360, liu2024infusion, mirzaei2024reffusion, hu2024innout, wu2025aurafusion360, haque2023instruct,  chen2024gaussianeditor} such as 3D scene editing or inpainting are applicable. LangSplat~\cite{langsplat} introduces language-aligned features into 3D Gaussians by distilling CLIP embeddings and multi-scale SAM~\cite{sam} masks from multi-view images, enabling open-vocabulary 3D segmentation. OpenGaussian~\cite{wu2024opengaussian} and Dr.Splat~\cite{jun2025drsplat} further enhance this pipeline by improving the quality of language features in Gaussian primitives through codebooks. ReferSplat~\cite{refersplat} tackles the challenging 3D referring expression segmentation (RES) task, aligning Gaussian features with sentence-level embeddings and predicting object masks conditioned on natural-language sentence queries. Despite promising results, these methods rely heavily on text-encoder embedding spaces and additional human annotation of ``sentence-object mask'' pairs on the training images, which limit their practicability on new scenes and ability to interpret nuanced or arbitrarily phrased queries.

\smallskip
In contrast, our method avoids the requirements of human-annotated description-mask pairs or embedding alignment entirely. Instead of training 3D features to match a fixed latent space, we generate rich captions for 3D instances and perform direct text-to-text retrieval using MLLMs, enabling more flexible reasoning over complex queries.
\section{Preliminary}
\label{sec:preliminary}


\begin{figure*}[!t]
    \centering
    \includegraphics[width=1.0\linewidth,]{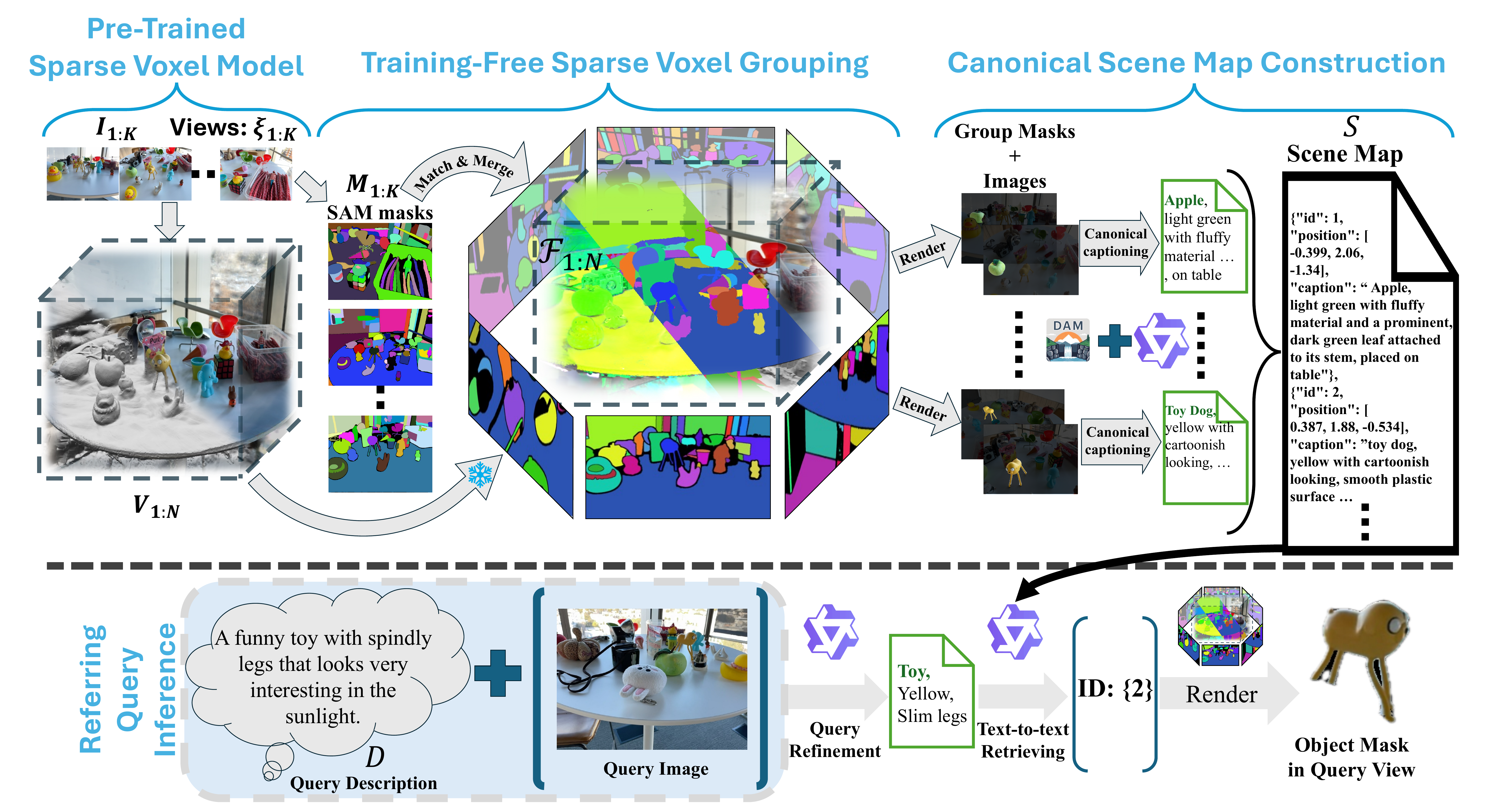}
    \caption{\textbf{Overview of \nickname.} Taking a sparse voxel model $V_{1:N}$ pre-trained from multi-view images $I_{1:K}$ and their corresponding camera pose $\xi_{1:K}$, we aim to build a voxel group field $\mathcal{F}_{1:N}$ from segmentation masks $M_{1:K}$ obtained from SAM2. With $\mathcal{F}_{1:N}$ we render images and masks for all groups to obtain canonical captions and construct a Scene Map $S$ recoding their position and captions. In the Referring Query Inference stage we take a query description and image to find the ideal target group that matches the description by query refinement and text-to-text retrieving from $S$, enabling complex segmentation tasks such as referring expression segmentation (RES).}
    \label{fig:overview}
\end{figure*}

We start by reconstructing scenes from multi-view input images. To this end, we employ the recently emerged SVR~\cite{svr}, which reconstructs scenes as sparse voxel fields and achieves high-quality reconstruction and real-time rendering. The reconstructed voxels are also suitable for efficiently assigning attributes to certain 3D locations, which our algorithm exploits. Their rendering follows the classical volume rendering equation, which is also the one employed by NeRF~\cite{mildenhall2021nerf} and 3DGS~\cite{kerbl202333dgs}:
\begin{equation} \label{eq:alpha_blending}
    \mathbf{C}(\mathbf{r}) = \sum_{i<N} w_i(\mathbf{r}) \cdot c_i, \quad w_i(\mathbf{r}) = \alpha_i(\mathbf{r}) \cdot \prod_{j<i} (1-\alpha_j(\mathbf{r})) \,,
\end{equation}
where $\mathbf{C}(\mathbf{r})$ is the rendering result for ray $\mathbf{r}$, $i$ goes through an ordered list of $N$ voxels in front of the ray, $c_i$ is the voxel attribute (usually color) to render, $\alpha_i(\mathbf{r})$ is the opacity of the ray passing through the voxel, and finally $w_i(\mathbf{r})$  is the contributing weight from the voxel to the rendered ray, which is also used in our algorithm during grouping.


\section{Methodology}
\label{sec:methodology}

We first define the notation and problem statement of our framework. As shown in Fig.~\ref{fig:overview}, given a 3D scene reconstructed as $N$ sparse voxels $\{V_i\}_{i=1}^N$ from a set of $K$ multi-view images $\{I_i \in \mathbb{R}^{H\times W\times 3}\}_{i=1}^K$ with their camera poses $\{\xi_i \in \mathrm{SE}(3)\}_{i=1}^K$, our goal is to assign descriptive language information to these voxels to construct a scene map $S$ and enable open-vocabulary reasoning from a natural language description $D$, which can be a vocabulary as in 3D-OVS~\cite{3dovs} or a referring expression as in 3D-RES~\cite{refersplat}. We target gradient-descent-free and latent-dimension-free approaches to save space and time.

To this end, we present \nickname, which consists of three main components: (1) \textbf{Training-Free Sparse Voxel Grouping} that groups voxels into coherent 3D instances (\cref{subsec:sparse_voxel_grouping}), (2) \textbf{Canonical Scene Map Construction} that stores the detailed captions and 3D spatial information for all voxel groups (\cref{subsec:grouped_voxel_captioning}), and (3) \textbf{Referring Query Inference} to support querying and reasoning with natural language input from users (\cref{subsec:text_to_text_reasoning}).

\subsection{Training-Free Sparse Voxel Grouping}
\label{subsec:sparse_voxel_grouping}

\begin{figure}[!t]
    \centering
    \includegraphics[width=0.8\linewidth,]{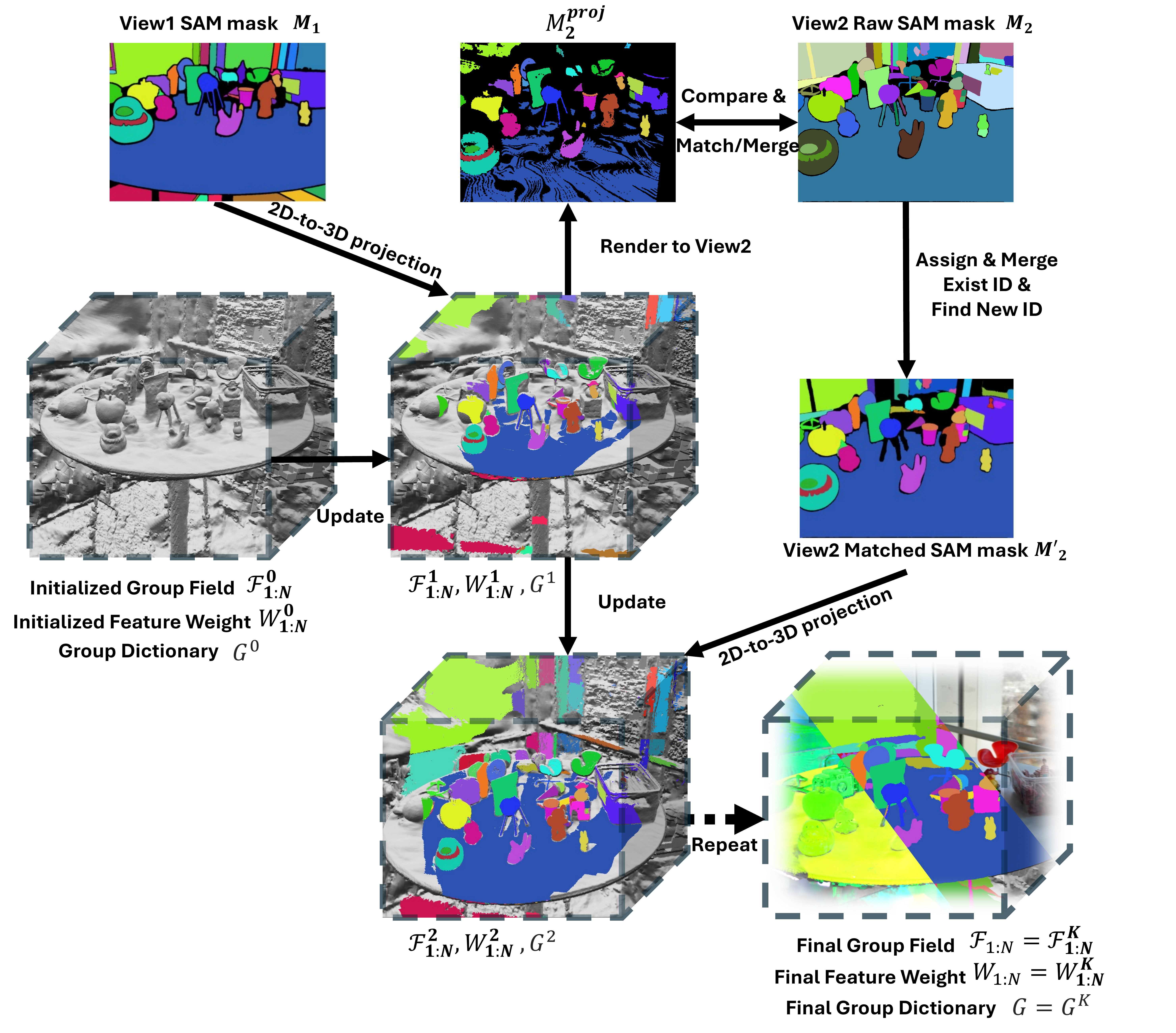}
    \vspace{-4mm}
    \caption{\textbf{Detail of the grouping process.} Taking the pre-trained voxel model $V_{1:K}$, we initialize the Group Field $\mathcal{F}^0_{1:N}$, Feature weight $W_{1:N}^0$ as empty tensors, and Group Dictionary $G^0$ as empty dictionary. Then start from $\xi_{1}$, we project the SAM masks $M_1$ to 3D voxel and update $\mathcal{F}_{1:N}$, $W_{1:N}$, and $G$. By match and merge masks from the other views repeating this process, the final $\mathcal{F}_{1:N}$, $W_{1:N}$, and $G$ is able to represent the group information of $V_{1:N}$.}
    \label{fig:grouping}
\end{figure}


Drawing inspiration from deep hough voting~\cite{deephoughvote} and spatial embedding~\cite{deepspatialclustering}, our key insight is that voxels belonging to the same object instance should cluster around a common 3D center position.
With this motivation, we encode group membership of voxels as 3D spatial coordinates pointing to their instance centroids.
These spatial embedding, termed as \textit{group feature}, is progressively updated by readily available 2D segmentation masks from foundation models like SAM2~\cite{ravi2024sam2}.
Our update is efficiently done in one training-free pass through all $K$ views without the lengthy gradient descent.
In contrast, existing methods like Gaussian Grouping~\cite{ye2023gaussiangrouping} rely on gradient descent to update per-primitive high-dimensional features with sequential training processes.

\paragraph{Group Representations.}
As depicted in Fig.~\ref{fig:grouping}, given a reconstructed scene of $N$ voxels, we extend voxel attributes by a group feature $F_{1:N} \in \mathbb{R}^{N\times 3}$, which denotes the instance centroid the voxels belonging to, and a feature weights $W_{1:N} \in \mathbb{R}^N$ indicating the confidence about the current feature.
We also use a Group Dictionary $G$ to record the spatial centroid of each unique instance.
The group features, weights, and dictionary are iteratively updated by the instance information from each frame.

\paragraph{Lifting 2D Segmentation Maps to 3D Group Field.}
We first introduce our method to lift a 2D instances mask into 3D voxels.
Let $M$ be the mask with $m$ unique instances ID, and $\mathbf{f}^{\mathrm{pts}} \in \mathbb{R}^{HW \times 3}$ be the rendered point map of the view, which indicates the expected ray-hit position in the world space.
The centroid $\mathbf{f}^{\mathrm{center}}_k\in\mathbb{R}^3$ of each instance $k$ is then determined by performing masked average reduction on the point map:
\begin{equation}
    \mathbf{f}^{\mathrm{center}}_k = \frac{\sum_{j \in HW} \mathbbm{1}(M_j = k) \cdot \mathbf{f}^{\mathrm{pts}}_j}{\sum_{j \in HW} \mathbbm{1}(M_j = k)} ~.
\end{equation}
We then accumulate the instance centroids information into each voxel $i$ by:
\begin{equation}
\label{eq:feat_update}
\begin{aligned}
    \mathcal{F}^{t+1}_{i} &= \mathcal{F}^t_{i} + \sum_{j \in HW} w_{ij} \,\mathbf{f}^{\mathrm{center}}_{M_j},\\
    W^{t+1}_i &= W^t_i + \sum_{j \in HW} w_{ij},
\end{aligned}
\end{equation}
where $w_{ij}$ is the blending weight (\cref{eq:alpha_blending}) the voxel $i$ contributing to the pixel $j$.
In practice, we customize the sparse voxel renderer~\cite{svr} to perform the update in one rendering pass.
The update equation follows Dr.Splat~\cite{jun2025drsplat}, while we do not need the top-k sampling as our group feature is only 3-dimensional.

\paragraph{Start condition.}
At view $\xi_1$, the voxel group feature $F^1$ and weight $W^1$ are directly assigned by the information lifted from the view.
The group dictionary is set to the $m$ instance IDs and their corresponding centroids $\mathbf{f}^{\mathrm{center}}$. 

\paragraph{Progressively Matching and Merging Segmentation Masks.}

When we move on to view $t+1$, we need to first match the mask IDs in $M^{t+1}$ with the existing voxel grouping $(F^t, W^t, G^t)$.
To this end, we render our group field into view $t+1$ as $M^{proj}_{t+1}$ and match existing instances with those in $M^{t+1}$.
The matched instance IDs are replaced with the existing IDs, while the remaining are added as new instances.
Specifically, we first determine the instance ID of each voxel $i$ in the current iteration by:
\begin{equation}
\label{eq:find_id}
    \mathrm{ID}_i^t = \operatorname{argmin}_j \left\lVert\frac{\mathcal{F}_i^t}{W_i^t} - G_j^t\right\rVert_2 ~,
\end{equation}
where the first term is the normalized voxel voting and the second term is the centroid of each instance.
We then cast rays from the processing view to retrieve the instance ID of the hit voxel, which forms a instance mask $M^{\mathrm{proj}}$ reflecting the existing segmentation.
Then, we conduct matching by finding the \textbf{\textit{highest}} IoU mask in $M^{t+1}$ for each instance in $M^{\mathrm{proj}}$.
We also prompt the SAM2 model again using $M^{\mathrm{proj}}$ and merge the instance masks that are largely overlapped to reduce noise and prevent the same instance from being incorrectly separated as two different IDs. 
As for the masks in $M^{t+1}$ that are neither matched nor overlapped with the existing ones, we assign a new unique ID and add them to the group dictionary.
Finally, we use the updated mask from $M^{t+1}$ to update the voxel group feature into $(F^{t+1}, W^{t+1})$ by \cref{eq:feat_update}.
After processing all the $K$ views, we use the $ID^K$ from \cref{eq:find_id} as the final voxel instance grouping.

It is worth noting that, with our Group Field construction strategy, even if the voxels are assigned to an incorrect group in some view, the accumulated $F_{1:N}$ and $W_{1:N}$ still force the voxel to ``\textit{vote}'' the most confident group it belongs, reducing the possible miss-grouping.

\subsection{Canonical Scene Map Construction}
\label{subsec:grouped_voxel_captioning}
\begin{figure}[!t]
    \centering
    \includegraphics[width=0.8\linewidth,]{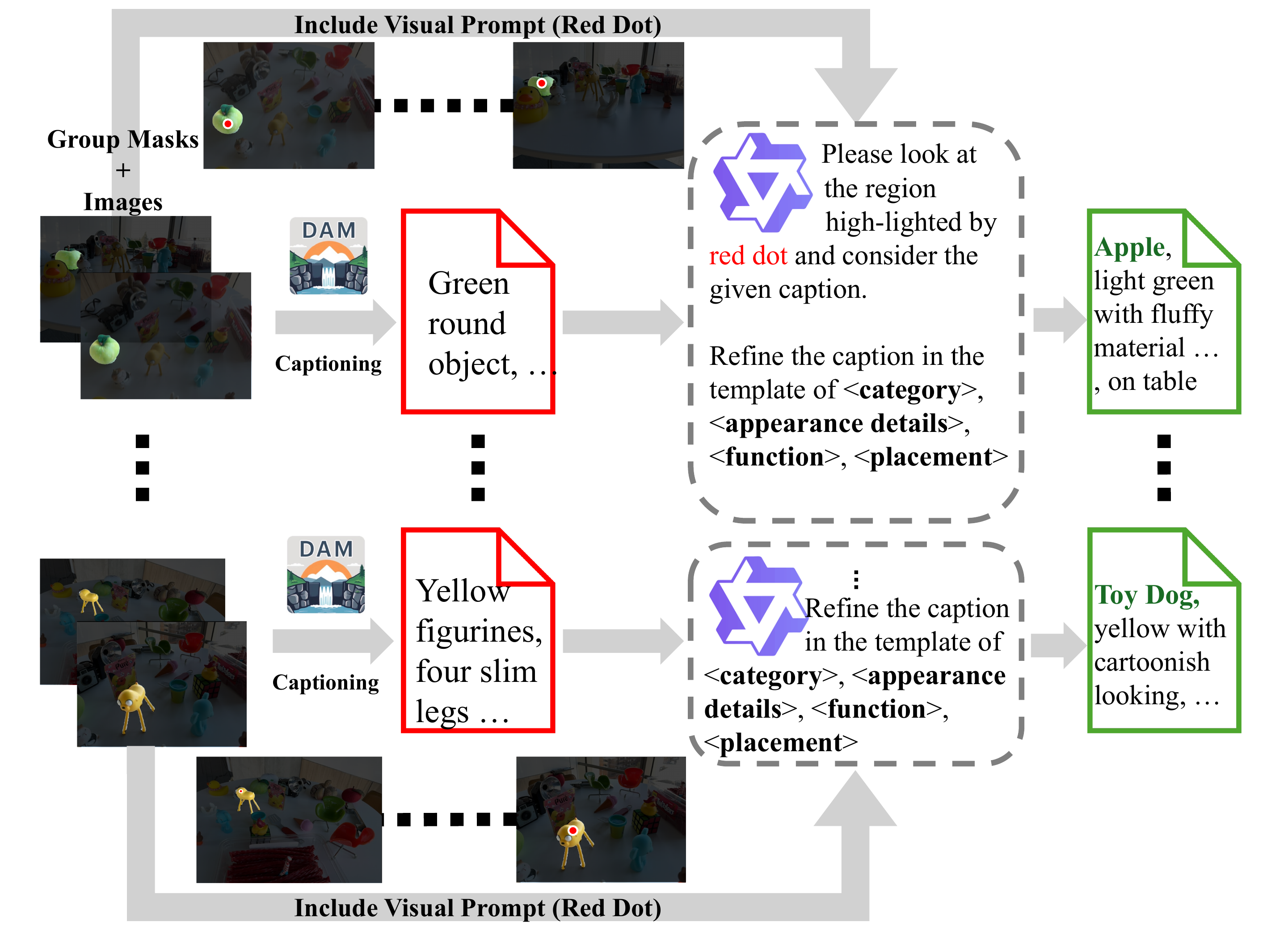}
    \vspace{-4mm}
    \caption{\textbf{Detail of the Canonical Captioning.} Given the group masks rendered of a specific group (taking the green apple as example) from our group field and their corresponding images, we leverage the Describe Anything Model (DAM) to first obtain a detailed caption. Then a Qwen3-VL model is conducted to canonicalize the caption into a fixed form, benefiting further usage.}
    \label{fig:captioning}
    \vspace{-2mm}
\end{figure}

Once voxel groups are obtained, our next goal is to construct a canonical scene map $S$ that records the position of each group and describes each group with a rich, human-readable caption, enabling open-vocabulary understanding of the entire scene (Fig.~\ref{fig:overview}). Instead of training 3D features to align with language embeddings, we bypass this bottleneck by leveraging Multimodal Large Language Models (MLLMs) to obtain detailed and accurate \emph{canonical} captions for each group. We now detail the canonical captioning process.

As depicted in Fig.~\ref{fig:overview} and Fig.~\ref{fig:captioning}, taking the green apple as an example, we render the binary masks of the corresponding group in views $\xi_{1:K}$ and use them as mask inputs together with the original images $I_{1:K}$ for the Describe Anything Model (DAM)~\cite{lian2025dam}. DAM produces a detailed description of the content inside each mask; however, its output is a free-form sentence and, in some cases, uses the generic word \emph{“object”} as the subject (e.g., “A green round object, possibly an apple, …”). To stabilize and standardize the captions across groups, we further use an MLLM (e.g., Qwen3-VL) to \emph{refine} these free-form captions into a fixed-form caption while considering (i) the DAM sentence, and (ii) the original masked images.

Because general MLLMs are not specifically trained for mask-image captioning, we design the visual prompting strategy inspired by~\cite {wu2024visualprompt}: we darken the regions outside the mask and highlight the object with a small red dot to focus the model’s attention. We then instruct the MLLM to rewrite the caption into the canonical template
\textit{\textless category noun\textgreater, \textless appearance details\textgreater\ \textless function/affordance or part-of\textgreater\ \textless placement/relation\textgreater}.
This converts each group’s description into a consistent, structured form. The resulting entry for a group in $S$ stores its ID, 3D position (center), and the canonical caption. In practice, the canonicalization significantly reduces subject ambiguity (e.g., replacing “object” with “apple”) and yields stable, comparable descriptions across views.

\subsection{Referring Query Inference}
\label{subsec:text_to_text_reasoning}



Given a user query $D$—from a single category token (e.g., “chair”) to a detailed referring expression (e.g., “what object can be used for cutting paper?”)—and an optional query image of the target view, our goal is to identify the best-matching 3D instance in the scene map $S$, obtain its ID(s), and render the corresponding binary mask in the target view. Rather than mapping the query and all captions into a learned embedding space, we perform \emph{direct retrieval} over $S$ with a multimodal LLM (MLLM). This design avoids any additional training or calibration for our 3D setting, maintains robustness to free-form phrasing that includes attributes, affordances, or light reasoning, and yields interpretable evidence because the selection is made over the canonical captions stored in $S$.

\paragraph{Query Refinement.}
To make matching deterministic and stable, we first \emph{canonicalize} the query. Following the same procedure as in our captioning stage, the MLLM rewrites $D$ (with or without the query image) into our fixed template \textit{\textless category noun\textgreater, \textless appearance details\textgreater\ \textless function/affordance or part-of\textgreater\ \textless placement/relation\textgreater}. For instance, the free-form description “A funny toy with spindly legs that looks very interesting in the sunlight” becomes “Toy, yellow, slim legs.” The refined query now matches the format of entries in $S$, so the retrieval process reduces to selecting the caption(s) in $S$ that best satisfy the refined description and returning their associated ID (In practice, we input the entire $S$ for the MLLM, so even if the query implies spatial relations (e.g., “left of the apple”), the MLLM can still possibly verify them using the stored group centers in $S$ before finalizing the selection.
\begin{table*}[t]
\centering
 \caption{\textbf{Quantitative evaluation on the Ref-LeRF subset in terms of referring expression segmentation (RES) mIoU.} ReferSplat* denotes the best results we reproduced from the official implementation. Note that only ReferSplat requires ground truth annotation of description-mask pair for training.}
\label{tab:reflerf_main}
\resizebox{0.9\textwidth}{!}{%
\begin{tabular}{lcccccc}
\toprule
\textbf{Method} & \textbf{Requires GT Annotation} &\textbf{ramen} & \textbf{figurines} & \textbf{teatime} & \textbf{kitchen} & \textbf{avg.} \\
\midrule
Grounded SAM~\cite{ren2024groundedsam} & -     & 14.1 & 16.0 & 16.9 & 16.2 & 15.8 \\
LangSplat~\cite{langsplat}         & - &12.0 & 17.9 &  7.6 & 17.9 & 13.9 \\
SPIn-NeRF~\cite{mirzaei2023spin}         & - & 7.3 &  9.7 & 11.7 & 10.3 &  9.8 \\
GS-Grouping~\cite{ye2023gaussiangrouping}       & - & 27.9 &  8.6 & 14.8 &  6.3 & 14.4 \\
GOI~\cite{qu2024goi}               & - & 27.1 & 16.5 & 22.9 & 15.7 & 20.5 \\
ReferSplat*~\cite{refersplat}       &\checkmark& 31.0 & 20.0 & 25.4 & 21.4 & 24.5 \\
ReferSplat~\cite{refersplat}        &\checkmark& 35.2 & 25.7 & 31.3 & 24.4 & 29.2 \\
\midrule
\textbf{OpenVoxel (Ours)} & - & \textbf{52.5} & \textbf{43.5} & \textbf{48.4} & \textbf{25.1} & \textbf{42.4} \\
\bottomrule
\end{tabular}}
\end{table*}


\paragraph{Text-to-text Retrieving of Target Group.}
Finally, given the selected ID(s), we render the referred object by rasterizing only the corresponding group(s) under the target view, producing a binary mask and the linked canonical caption from $S$. This inference procedure is fully training-free, applies uniformly to both OVS (category-only queries) and RES (attribute/affordance/relation queries), and empirically provides stable matches across scenes and phrasing styles while remaining simple to implement.

\section{Experiments}
\label{sec:exp}
\subsection{Settings}
\label{subsec:setting}
\paragraph{Datasets.}
Following prior state-of-the-art works~\cite{langsplat, lerf, refersplat}, we evaluate the open-vocabulary segmentation (OVS) and referring expression segmentation (RES) performance of our \nickname\ primarily on subsets of the iPhone Polycam-captured LeRF dataset~\cite{lerf}. 
For the OVS task, we follow the evaluation protocols in~\cite{ye2023gaussiangrouping, zhu2025objectgs} using the \textit{LeRF-Mask} subset, which contains three scenes (\textit{ramen}, \textit{figurines}, and \textit{teatime}) with 6, 7, and 10 queryable objects, respectively. We further conduct experiments on the \textit{LeRF-OVS} subset~\cite{langsplat}, with the same three scenes with 13, 17, and 12 annotated objects, respectively.
For the RES task, we use the \textit{Ref-LeRF} subset~\cite{refersplat}, which consists of four scenes (\textit{ramen}, \textit{figurines}, \textit{teatime}, and \textit{kitchen}) containing 13, 17, 12, and 11 objects, respectively, each paired with sentence-level natural language referring expressions.

\paragraph{Implementation details.}
We conduct our \nickname with PyTorch~\cite{paszke2019pytorch} libraries, where our code base is built on top of SVR~\cite{svr}. For the foundation models, we use SAM2~\cite{ravi2024sam2} to obtain the per-view segmentation map separately and also prompt SAM2 to conduct the mask merging operation during the Training-Free Sparse Voxel Grouping, as described in Sect.~\ref{subsec:sparse_voxel_grouping}. On the other hand, the Describe Anything Model(DAM)~\cite{lian2025dam} is used as our captioning model. 
As for the MLLM used to refine the obtained captions/query and conduct the target retrieving, we use Qwen3-VL-8B-Instruct~\cite{Qwen-VL, Qwen2-VL, Qwen2.5-VL}. 
Please refer to our supplementary material for more implementation details, including engineering tricks and the system prompt we use.

\subsection{Quantitative Evaluations}
\label{subsec:quantitative}

\begin{table}[t]
\centering
 \caption{\textbf{Quantitative evaluation on the LeRF-OVS subset in terms of mIoU.} }
\label{tab:lerfovs_main}
\resizebox{0.7\linewidth}{!}{%
\begin{tabular}{l|ccc|c}
\toprule
\textbf{Method} & \textbf{ramen} & \textbf{figurines} & \textbf{teatime} & \textbf{avg.} \\
\midrule
Feature-3DGS~\cite{zhou2024featuregs}      & 43.7 & 58.8 & 40.5 & 47.7 \\
LEGaussians~\cite{shi2024legaussian}       & 46.0 & 60.3 & 40.8 & 49.0 \\
LangSplat~\cite{langsplat}         & 51.2 & 65.1 & 44.7 & 53.7 \\
GS-Grouping~\cite{ye2023gaussiangrouping}       & 45.5 & 60.9 & 40.0 & 48.8 \\
GOI~\cite{qu2024goi}               & 52.6 & 63.7 & 44.5 & 53.6 \\
3DVLGS~\cite{peng20243dvlgs}            & 61.4 & 73.5 & 58.1 & 64.3 \\
ReferSplat~\cite{refersplat}        & 55.1 & 67.5 & 50.1 & 57.6 \\
CCL-LGS~\cite{tian2025ccllgs}           & 62.3 & 61.2 & 71.8 & 65.1 \\
\midrule
\textbf{OpenVoxel (Ours)} & \textbf{62.5} & \textbf{60.7} & \textbf{75.4} &  \textbf{66.2} \\
\bottomrule
\end{tabular}}
\end{table}
\begin{figure*}[!t]
    \centering
    \includegraphics[width=0.9\textwidth,]{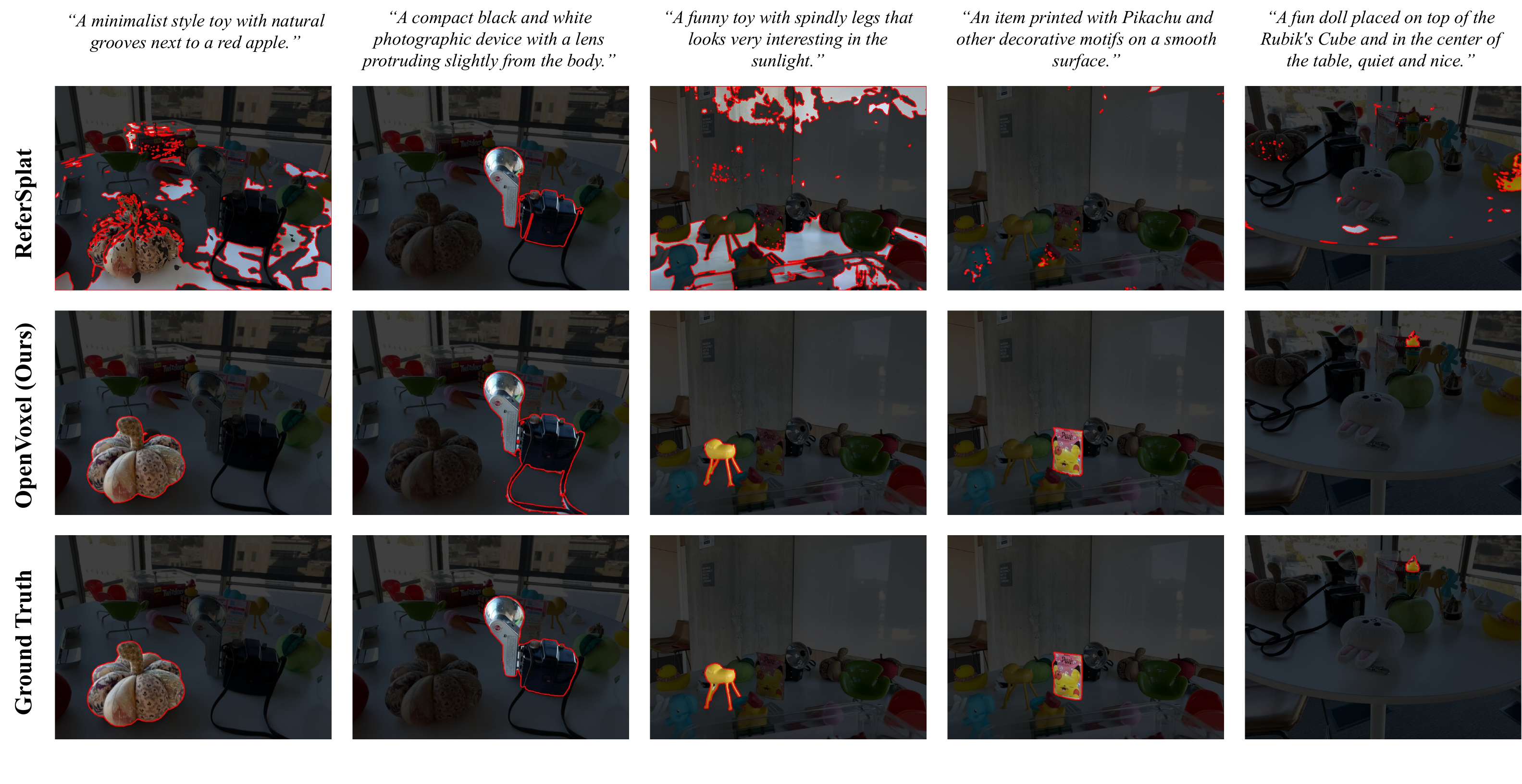}
    \vspace{-4mm}
    \caption{\textbf{Qualitative results of RES task on the \textit{Figurines} scene.}}
    \label{fig:ref_figurines}
\end{figure*}

\begin{table*}[ht]
\centering
 \caption{\textbf{Quantitative evaluation on the LeRF-mask subset in terms of IoU and BIoU.} Note that BIoU indicates the boundary IoU of predicted mask and ground-truth mask. Best results are in bold text, and second best results are underlined.}
\label{tab:lerf_mask_main}
\resizebox{0.9\textwidth}{!}{%
\begin{tabular}{l| cc cc cc |cc}
\toprule
\multirow{2}{*}{\textbf{Model}} & \multicolumn{2}{c}{\textbf{figurines}} & \multicolumn{2}{c}{\textbf{ramen}} & \multicolumn{2}{c|}{\textbf{teatime}} & \multicolumn{2}{c}{\textbf{Avg.}}\\
\cmidrule(lr){2-3}\cmidrule(lr){4-5}\cmidrule(lr){6-7}\cmidrule(lr){8-9}
 & mIoU & mBIoU & mIoU & mBIoU & mIoU & mBIoU & mIoU & mBIoU\\
\midrule
DEVA~\cite{cheng2023deva}           & 46.2 & 45.1 & 56.8 & 51.1 & 54.3 & 52.2 & 52.4 & 49.5 \\
LERF~\cite{lerf}           & 33.5 & 30.6 & 28.3 & 14.7 & 49.7 & 42.6 & 37.1 & 29.3\\
SA3D~\cite{cen2023sa3d}           & 24.9 & 23.8 &  7.4 &  7.0 & 42.5 & 39.2 & 24.9 & 23.3\\
LangSplat~\cite{langsplat}      & 52.8 & 50.5 & 50.4 & 44.7 & 69.5 & 65.6 & 61.2 & 56.1\\
GS Grouping~\cite{ye2023gaussiangrouping}    & 69.7 & 67.9 & 77.0 & 68.7 & 71.7 & 66.1 & 72.8 & 67.6\\
Gaga~\cite{lyu2024gaga}           & \underline{90.7} & \textbf{89.0} & 64.1 & 61.6 & 69.3 & 66.0 & 78.5 & 74.2\\
ObjectGS~\cite{zhu2025objectgs}       & 88.2 & 85.2 & \textbf{88.0} & \textbf{79.9} & \textbf{88.9} & \textbf{88.6} & \textbf{88.3} & \textbf{84.4}\\
OpenVoxel (Ours) & \textbf{90.8} & \underline{89.0} & \underline{84.9} & \underline{73.2} & \underline{85.8} & \underline{82.1} & \underline{87.2} & \underline{81.4} \\
\bottomrule
\end{tabular}}
\end{table*}

\paragraph{Ref-LeRF.}
We first present the RES results on the Ref-LeRF subset in Table~\ref{tab:reflerf_main}, as this task best demonstrates the advantages of our design. 
Since the official pre-trained models and data of the current state-of-the-art method, ReferSplat~\cite{refersplat}, do not cover all four scenes, we reproduce its results following the original implementation and configurations, denoted as ReferSplat*. 
As shown in Table~\ref{tab:reflerf_main}, despite \nickname\ not requiring any description–mask annotations during grouping or scene map construction, it achieves a notable improvement of \textbf{13.2\%} over the result reported in the original ReferSplat paper, and an even larger margin of \textbf{17.9\%} compared to our reproduced version. 
During reproduction, we observe that ReferSplat’s strategy of learning sentence-level embeddings for 3D representations tends to overfit the seen descriptions, resulting in unstable evaluation performance. 
In contrast, our \nickname\ consistently achieves higher accuracy, and the superior performance of our text-level retrieval mechanism supports its effectiveness and robustness compared to learned embedding approaches.

\paragraph{LeRF-OVS and LeRF-Mask.}
Tables~\ref{tab:lerfovs_main} and~\ref{tab:lerf_mask_main} summarize the OVS results on the LeRF-OVS and LeRF-Mask subsets, respectively. 
For LeRF-OVS, we report mean Intersection-over-Union (mIoU) between predicted and ground-truth masks, following~\cite{langsplat, refersplat}. 
For LeRF-Mask, we additionally evaluate mean Balanced IoU (mBIoU), following~\cite{ye2023gaussiangrouping, zhu2025objectgs}. 
As shown in Table~\ref{tab:lerfovs_main}, the OVS task involves relatively simpler queries, leading to higher overall performance across methods—even when evaluated on the same target objects as in Ref-LeRF. 
For LeRF-Mask, where the queries are fewer and less ambiguous, all current state-of-the-art methods (including ours) achieve over 70\% mIoU. 
Although our design primarily targets the RES setting, \nickname\ still performs competitively on both OVS benchmarks, demonstrating its flexibility and robustness across varying query complexities.

\begin{figure*}[!t]
    \centering
    \includegraphics[width=0.9\textwidth,]{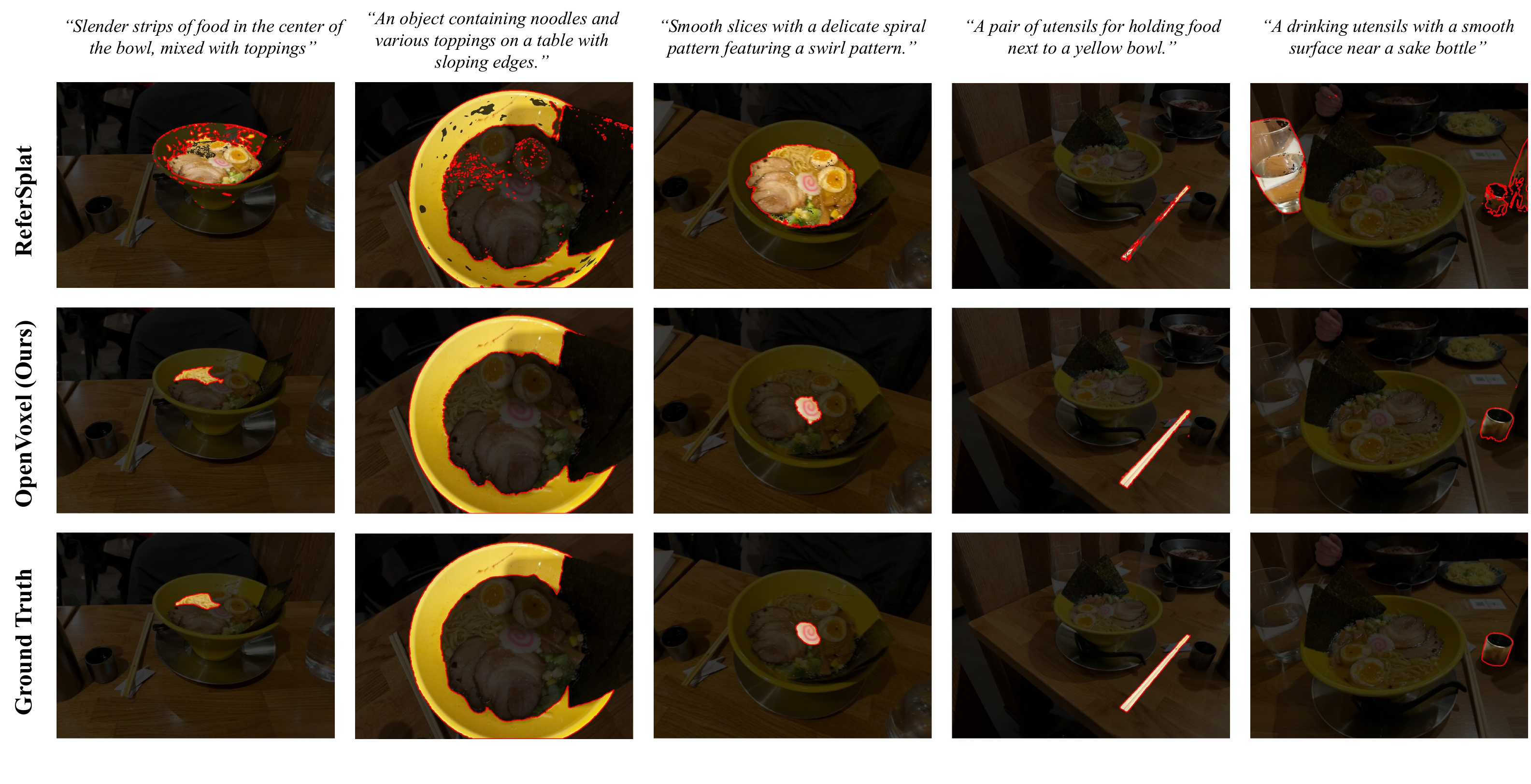}
    \vspace{-4mm}
    \caption{\textbf{Qualitative results of RES task on the \textit{Ramen} scene.}}
    \label{fig:res_ramen}
    \vspace{-4mm}
\end{figure*}
\subsection{Qualitative Results}
\label{subsec:qualitative}
We show the qualitative results of \textit{``figurines''} and \textit{``ramen''} scene in Ref-LeRF subset in Fig.~\ref{fig:ref_figurines} and Fig.~\ref{fig:res_ramen}, respectively. In both scenes, ReferSplat often identifies only partial regions of the target object. For example, in Fig~\ref{fig:ref_figurines}, when we input the query \textit{``A minimalist style toy with natural
grooves next to a red apple''}, ReferSplat focuses on the region near the apple (the apple is at the left of the image), without localizing the correct object. In contrast, \nickname retrieves the toy pumpkin properly and render a smooth mask. As for the query \textit{``A drinking utensils with a smooth
surface near a sake bottle''} in Fig.~\ref{fig:res_ramen}, ReferSplat only focus on the term \textit{``drinking utensils''} and incorrectly segments both the glass of water and the sake bottle, while our \nickname locates the sake cup correctly. These results further support our claim of using explicit captions to build a scene map and perform text-level retrieval, which benefits the RES tasks.

\subsection{Further Analysis}
\label{subsec:analysis}
\paragraph{Ablation Study.}

Table~\ref{tab:ablation} shows the ablation studies of our \nickname on the Ref-LeRF subset. Start from the baseline \textbf{A} where we do not merge overlapping small masks as described in Sect.~\ref{subsec:sparse_voxel_grouping} during the voxel grouping, and also do not canonicalize both the caption and the query. We can see that even with the explicit captioning and text-to-text retrieval, it is still hard to retrieve the correct target due to noisy small groups and misalignment of language format. After that, we include the mask merging strategy in model \textbf{B}, showing a slight improvement of $3.7$\% mIoU due to less noisy groups. Then, in model \textbf{C}, we introduce the canonical captioning for each group while constructing the scene map, showing a large improvement of $8.4$\% mIoU because of more specific captioning (i.e., standard template for each caption and reducing the word ``object'' as subject noun). Lastly, our full model canonicalizes the input text query into the same form as the canonical caption, introducing more precise alignment and less ambiguity, and showing the best performance. The above ablation study verifies the necessity of each of our designs.  
\begin{table}[t]
\centering
 \caption{\textbf{Ablation study of our \nickname on Ref-LeRF subset.} We integrative evaluate the effectiveness of several components of our \nickname.}
\label{tab:ablation}
\resizebox{0.7\linewidth}{!}{%
\begin{tabular}{l|ccc|c}
\toprule
\textbf{Method} & \textbf{Mask Merging} &\textbf{Canonical Caption} & \textbf{Canonical Query} & \textbf{mIoU} \\
\midrule
A & -     & - & - & 24.3 \\
B & \checkmark     &-  & -&  28.0\\
C & \checkmark     & \checkmark & - &  36.4 \\
\midrule
\textbf{Ours} & \checkmark & \checkmark & \checkmark & \textbf{42.4} \\
\bottomrule
\end{tabular}}
\end{table}
\paragraph{Runtime Estimation.}

We show the comparison of run time between our \nickname and current SOTAs on OVS and RES tasks in Table~\ref{tab:run_time} on a single RTX 5090 GPU. The term ``Requires Training'' here indicates if a method needs training to obtain its feature/language field. Although ObjectGS~\cite{zhu2025objectgs} combines the training process of 3DGS and the per-Gaussian object feature (similar to Gaussian Grouping), we note that running a vanilla 3DGS on RTX 5090 only takes about 10 minutes, indicating ObjectGS needs an additional 30 minutes to obtain the object feature. On the other hand, although ReferSplat reports that it only takes 58 minutes to train on an A6000 GPU, in all our reproductions, we find that following the official implementation configuration takes at least 2 hours per scene to obtain the best results. In contrast, our \nickname requires only about 3 minutes to complete the voxel grouping and canonical scene map construction, which is at least 10 $\times$ faster than other SOTAs, and the inference time per query is less than 1 sec.
\begin{table}[t]
\centering
 \caption{\textbf{Run time comparison with current SOTAs.} For the term ``Requires Training'', it is to indicate if a method needs gradient-based training to obtain the semantic or language field. All methods are reproduced on a single RTX 5090 GPU.}
\label{tab:run_time}
\resizebox{0.65\linewidth}{!}{%
\begin{tabular}{lcc}
\toprule
\textbf{Method} & \textbf{Requires Training} &\textbf{Run Time} \\
\midrule
ReferSplat~\cite{refersplat} & yes & \textgreater 1 hr\\
ObjectGS~\cite{zhu2025objectgs} & yes & $\sim$ 40 min\\
\midrule
OpenVoxel (Ours) & no &  $\sim$ 3 min\\
\bottomrule
\end{tabular}}
\end{table}
\section{Conclusion}
\label{sec:conclusion}
In this paper, we propose \textit{\nickname}, a training-free framework for grouping and captioning sparse voxel rasterization (SVR) models. In our \textit{Training-Free Voxel Grouping} process, we curate a group field for a pre-trained SVR model by matching and merging per-view SAM2 masks. As for the \textit{Canonical Scene Map Construction} we leverage VLMs and MLLMs to obtain a fix form of caption for each constructed group and then collect positions and captions of all groups as the scene map. And finally, we conduct a \textit{Referring Query Inference} to retrieve target objects from our scene map with an input descriptive query. Through quantitative and qualitative experiments, we verify the efficiency and capability of \nickname.

\clearpage
\appendix
\section*{Appendix}
In this supplementary material, we provide additional method details in \cref{supp:details} and more results of semantic segmentation in \cref{supp:semseg}, ablation studies in \cref{supp:abla}, visualization in \cref{supp:qual}.
Finally, we discuss about our limitation in \cref{sec:limiation}.

\section{Details}
\label{supp:details}

\begin{figure}[!t]
    \centering
    \includegraphics[width=1.0\linewidth,]{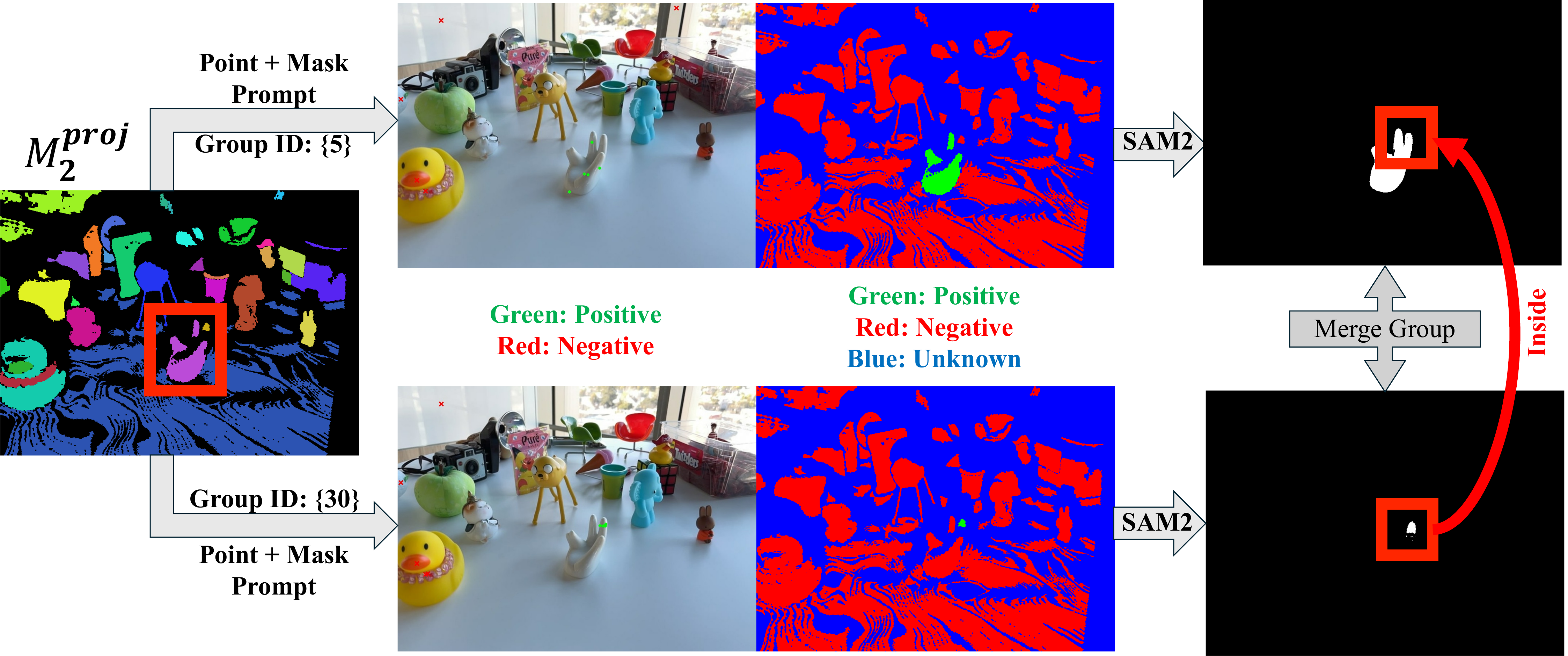}
    \caption{\textbf{Detail of the mask merging process.} }
    \label{fig:merging}
\end{figure}

\paragraph{Merging groups.}
In Sect.~\ref{subsec:sparse_voxel_grouping}, we mentioned merging small segmentation masks by re-prompting SAM2~\cite{ravi2024sam2} to reduce noise. Here, we visualize this process for a better understanding. As depicted in Fig.~\ref{fig:grouping} and Fig.~\ref{fig:merging}, the \textit{``hand sculpture''} instance is separated into two different groups in $M^{proj}_2$ since the SAM2 mask $M_1$ treats them as two different segment, where one group covers almost the entire hand and the other just representing a finger tip. To encourage each of our groups to be more instance-wise, we double-check if any two groups should be merged by prompting all groups that are observable under the current view with SAM2 again. Taking the hand sculpture under view $\xi_2$ as example in Fig.~\ref{fig:merging}, we create the point prompts for SAM2 by treating several pixels sampled from the group in $M^{proj}_2$ of interest as positive prompts, and also select several other groups and treat their center pixel in $M^{proj}_2$ as negative prompts. As for the mask prompt, we treat the entire group of interest in $M^{proj}_2$ as a positive prompt (mask value: $20$), all the other known groups as a negative prompt (mask value:$-20$), and the rest of the region is treated as unknown (mask value:$0$). After using this setting as prompt for SAM2, if the output mask of one group (e.g., the finger tip in Fig.~\ref{fig:merging}) is almost lie inside another group's mask (e.g., over $90 \%$ inside the hand sculpture), then we merge the smaller group into the larger group accordingly (i.e., merge the finger tip into the hand sculpture). As shown in our ablation study Table~\ref{tab:ablation}, a slight improvement is observed by including this merging strategy.  

\paragraph{MLLM prompt.}
In Sect.~\ref{subsec:grouped_voxel_captioning} and~\ref{subsec:text_to_text_reasoning}, we leverage an MLLM model of QWen3-VL-8B to conduct canonical captioning, prompt refinement, and target retrieving. The system prompts are as List~\ref{code:prompt_caption}, List~\ref{code:prompt_query}, and List~\ref{code:prompt_retrieve}.

We note that we do not spend much effort exploring different kinds of system prompt design. We simply use ChatGPT by describing our task with some examples and ask it to provide system prompts that suit Qwen-VL~\cite{Qwen2.5-VL, Qwen2-VL, Qwen-VL} models, and the same prompts are shared for both RES and OVS tasks. Therefore, these system prompts may not be the optimal ones, and users still have a chance to improve the performance by using our pipeline and simply changing system prompts. 

\paragraph{Implementation.}
We now elaborate on the implementation details. Since our method is totally training-free after having the pre-trained SVR model of each scene, reducing the process time of the following processes (i.e., grouping, captioning, and retrieving) is a crucial problem. To achieve this goal, we do not always go through all the images from the training set for each scene in practice. Instead, taking the LeRF~\cite{lerf} dataset and corresponding subsets as an example, we uniformly sample the processed views to make sure the total processed view of each scene does not exceed $150$. Also, we found that since the merging process requires additional SAM2 execution, conducting this merging process for all views would slow down the inference time. Hence, we conduct the merging (re-prompting SAM2) process for each scene per $1$ to $5$ steps to speed up the inference. Furthermore, when prompting the captioning model (i.e., DAM~\cite{lian2025dam}), we only sample  $8$ frame-mask pairs (pad to $8$ pairs if not enough) for each group to reduce the visual tokens for the model, making sure that the inference is fast. Also, for the usage of MLLM for Canonical Captioning, Query Refinement, and Text-to-text retrieving, we ``\textit{DO NOT}'' provide any visual example for the MLLM as in-context examples since the inference time would be slowed down by doing so (although we acknowledge that conducting these information might bring better performance, we leave this as a future direction to explore the balance between adding visual demonstrations and inference time).

\begin{promptlisting}[language=Python]
CANONICAL_CAPTION_SYSTEM_PROMPT= """You are a detail-focused visual caption 
refiner for open-vocabulary segmentation and referring grounding.
INPUTS
- A short video where ONE region is masked (highlighted); the outside area 
is darkened.
- A rough caption from another model (may be incorrect or misleading).
SCOPE
- Describe ONLY what lies INSIDE the masked region across frames.
- Never name or infer unmasked neighbors as the subject.
- Be factual; do not guess hidden details.
- If the original caption conflicts with the visual evidence, IGNORE it and 
correct the errors.
REWRITE GOAL
Produce a precise, natural description with a clear class noun and 
discriminative details that is easy to match with open-vocabulary queries.
CORE RULES
1) Class noun: Replace vague words ("object/thing/item/surface") with a 
concrete class or fine-grained subtype.
2) Part-of decision (strict):
   - Use "part of <larger object>" ONLY if ALL are true:
     a) Visible physical continuity/attachment within the mask (seam/stitch/
     joint/hinge/fastener or continuous material/geometry),
     b) The region is an intrinsic component,
     c) The larger object's category is visibly identifiable.
   - Otherwise DO NOT use "part of". Prefer placement instead.
3) Placement vs background:
   - Use view-independent placement for surfaces/containers ("on table", 
   "inside pouch", "on plate", "on shelf", "in tray", etc.).
   - If the region is background material/texture (floor/wall/ceiling/ground), 
   begin with "background: <material/surface>".
4) Printed-content rule (critical):
   - Scan ALL frames within the mask for printed text, logos, labels, 
   or characters.
   - If ANY are visible, include at least one cue:
     - Text: transcribe exact readable tokens (keep visible case/punctuation). 
     If partial, include the visible substring.
     - Character/graphic: if identity is uncertain, 
     describe visual attributes neutrally (e.g., "purple cartoon dinosaur"); 
     do not guess names.
5) Detail quota:
   - Include at least FOUR distinct cues chosen from: color; material; 
   texture/pattern; shape/geometry; subtype/model; visible text/logo/
   printed character; state/condition; function/affordance; 
   part-of (only if allowed); placement/relation (max 2).
6) Language hygiene:
   - View-independent wording only (no left/right/front/top; no camera terms).
   - Do not mention the highlight/red dot.
   - Forbidden words: object, thing, item (and similar generic fillers).
OUTPUT FORMAT (canonical; must be strictly followed)
- EXACTLY ONE line, 12-20 words, comma-separated phrases, no period.
- Start with the SUBJECT noun.
- **Order of phrases (strict):**
  1) <category noun> (table, chair, bottle, pouch, human character, cat, 
  dog, rabbit, camera, spoon, door handle, floor, wall, etc.)
  2) <appearance details>  (color/material/texture/pattern/shape/
  subtype/text/logo)
  3) <function/affordance or part-of>  ("part of <larger object>" 
  only if rule 2 allows; otherwise a concise function/affordance like 
  "resealable pouch", "pour spout", "grip handle")
  4) <placement/relation>  (on/in/inside/attached to/against/between;
  max 2 relations)
- If unsure between "part of" and placement, choose placement.
- For background regions: "background: <material/surface>, 
<appearance details>, <(optional) function if any>, <placement/relation>".
STRICTNESS
The form of the output must strictly follow the rules above and the 
ordered template:
<category noun (no color)>,(comma here)
<appearance details (color here)><function/affordance or part-of>
<placement/relation>.
The original caption may be wrong; rely on visual evidence to correct it.
"""
\end{promptlisting}
\captionof{listing}{System prompt of Canonical Captioning.}
\label{code:prompt_caption}

\begin{promptlisting}[language=Python]
QUERY_REPHRASE_SYSTEM_PROMPT = """You are a helpful assistant.
You rewrite a short OVS query into ONE short canonical phrase
using ONLY the provided image and the raw query.

SCOPE
- Inputs:
(a) scene_map (JSON list of candidate objects with their coordinates and caption),
(b) query (description about an object visible in the image),
(c) view_image.
- Keep the result SHORT and human-judgable from the query mainly.
- You must NOT include any spatial relations in the output if not explicitly
mentioned in the query.

CANONICAL FORM
- Output exactly ONE short phrase in the form:
<class noun> <appearance> <placement?>
- 2 to 6 words, lowercase, spaces only, no punctuation.
- class noun: singular, most specific common name that is visually supported
(e.g., "rubber duck", "paper bag").
- appearance: brief, image-supported attributes (color/material/texture/
state/text/logo/shape). If unsure, keep generic
(e.g., "plastic-like", "transparent").
- placement (OPTIONAL): view-INDEPENDENT, simple scene phrase
(e.g., "on table", "in bowl", "on shelf", "in bag").
Avoid left/right/front/behind/above/below.

REPHRASE PROTOCOL (follow strictly)
- Be conservative if uncertain; never hallucinate specifics you cannot see.
- Examples:
  - "toy car on the table" -> "car on table"
  - "banana" -> "banana" (no assumed color)
  - Materials: "plastic bag"->"plastic-like bag"; "nori"->"seaweed";
    "glass cup"->"transparent cup"; "porcelain"->"ceramic"
  - Common words: "gummy"->"gummy candy"; "ribeye beef"->"piece of meat";
    "toy car"->"car"; "stuffed bear"->"teddy bear";
    "paper napkin"->"napkin"; "kamaboko"->"small piece with pink swirl";
    "rubber duck with a bouy"->"rubber duck with pink lei"
  - Character names -> descriptions:
    "pikachu"->"yellow character with long ears and possibly red cheek";
    "jake"->"yellow cartoon character big eyes slim legs";
    "miffy"->"rabbit character, possibly wearing garment";
    "waldo"->"character red-white striped shirt";
    "hello kitty"->"white cartoon cat red bow"
  - Brands -> generic: "lays"->"potato chips"; "coca-cola"->"can";
    "nike shoes"->"sports shoes";
    "tesla door handle"->"metalic object, look like door handle"
  - Ambiguous placements: "in the bowl"/"on the plate"/"inside the pouch"
    -> "in container"/"on surface"/"in bag"

OUTPUT (STRICT)
Return ONE JSON line only (no extra text, no code fences, no reasoning):
{"canonical": "<clear class noun>, (you must include this comma after <clear class noun>) <appearance (color)>, <placement (ONLY IF contained in query text)>"}

CONSTRAINTS
- No chain-of-thought or explanations.
- Do not use any information that cannot plausibly be inferred from
the image + query alone.
- You MUST NOT use ambiguous noun like "object", "thing", "item", "stuff",
"part", "area", "region", "section", "portion", "background", "foreground",
"surface", "area of interest", etc.
- If the class noun is a general category
(e.g., "container", "food", "furniture"),
you MUST add more specific appearance to clarify.
- If the query does not contain any placement info,
DO NOT add any placement in the output.
"""
\end{promptlisting}

\captionof{listing}{System prompt of Query Refinement.}
\label{code:prompt_query}

\begin{promptlisting}[language=Python]
SYSTEM_PROMPT_RETRIEVE = """You retrieve all matching targets using 
scene_map + view_image (optional) + a canonical short phrase.

INPUTS
1) scene_map: JSON of candidate voxel groups with fields:
   - id (integer, unique)
   - caption (short description; copy EXACTLY in output)
   - center: WORLD coordinates (use only for view-independent relations:
     near/far/between/closest/farthest)
   - optional: bbox/size/group/category
2) view_image (optional): one image for the current query instance
   (targets may be occluded or off-frame).
3) canonical: the short canonical phrase from Stage 1
   (e.g., "rubber duck, yellow", "paper bag, on table").

POLICY (caption-first, occlusion-robust)
- Primary signal: scene_map CAPTIONS (semantic match to the canonical phrase;
  allow common synonyms/hypernyms).
- Ignore view-dependent relations (left/right/front/behind).
  Use WORLD coords ONLY for near/far/between/closest/farthest
  if such words appear.
- One real object may be split across multiple voxel groups (ids)
  that are spatially adjacent and semantically consistent.
  If so, RETURN ALL ids for that instance.
- If multiple separate instances match the canonical phrase,
  RETURN the best aligned one.
- Secondary signal: view_image (if provided) only to veto
  obvious mismatches when visible; do NOT penalize occlusion.

INTERNAL STEPS (do not reveal):
1) Match captions to the canonical phrase -- prioritize
   exact/synonym class match, then attribute alignment.
2) Merge adjacent voxel groups that describe the same instance
   (spatially close in WORLD coordinates and semantically consistent).
3) If canonical phrase includes near/far/between/closest/farthest,
   apply these using WORLD centers/bboxes over the matched set.
4) Use the image (if provided) only to down-weight
   clear visual contradictions when visible (do not discard due to occlusion).
5) Finalize ids and copy their captions EXACTLY from scene_map.

OUTPUT (STRICT)
Return EXACTLY one JSON line -- no extra text,
no code fences, no reasoning:
{"ids": [<int>, ...], "captions": ["<EXACT caption>", ...]}
Rules:
- Include at least one id for every matching instance
  (multi-instance allowed, but in most case only one).
- Sort ids ascending within each instance; overall order is arbitrary.
- Captions must be copied EXACTLY from scene_map, same order as ids.
- You cannot return empty ids or ids that are not in scene_map.
- If borderline: you MAY add
  {"candidates": [{"id": a}, {"id": b}]}

CONSTRAINTS
- No chain-of-thought or explanations.
- Do not paraphrase any caption in the "captions" array;
  copy exactly from scene_map.
- Use WORLD coordinates only for view-independent relations;
  do not use image axes for left/right/front/behind.
"""
\end{promptlisting}

\captionof{listing}{System prompt of target retrieval.}
\label{code:prompt_retrieve}

\section{Semantic Segmentation} \label{supp:semseg}

\subsection{Approach}
\label{subsec:sem_seg_approach}
Different from OVS and RES, the task of semantic segmentation usually has a customized list of class candidates for each dataset. Therefore, instead of retrieving the matched groups for each class, we conduct semantic segmentation for \nickname by choosing the best-matched class for each group. 

\subsection{Dataset and implementation details}
\paragraph{Dataset.}
Following OpenGaussian~\cite{wu2024opengaussian}, we conduct semantic segmentation on $10$ different scenes on the Scannet~\cite{dai2017scannet} dataset. Each scene is represented as colored point clouds, with ground truth images and depth maps provided. Following the official setting, we conduct a $19$ class semantic segmentation in our experiments.

\paragraph{Implementation details.}
We note that in OpenGaussian, ground truth point clouds are directly utilized as initialization for the 3DGS, and they deactivate all the merging and splitting processes so that a perfect geometry alignment is naturally obtained for evaluation. However, it is not easy for our backbone (i.e., SVR~\cite{svr}) to have such an initialization. To have a better geometry for the Scannet dataset, we utilize the provided depth map to guide the pre-training process of the SVR model.

\subsection{Evaluation protocols and results}
\paragraph{Evaluation protocols.}
 Since we do not use the ground truth points for pre-training the SVR model, the constructed voxel number of each scene is very different from the number of ground truth points. Typically, the number of ground truth point clouds is about $50$K to $350$K, but the number of our voxels are about $5$M to $10$M. Therefore, we conduct several different protocols for evaluations to better showcase our \nickname:
 \textbf{(1) Nearest},\textbf{(2) Majority of $25$-NN}, and \textbf{(3) Majority of $50$-NN}. The \textbf{Nearest} protocol means that for each point in the ground truth point cloud, we find the spatially nearest voxel and take the voxel's class ID (obtained as described in~\ref{subsec:sem_seg_approach}) as our prediction; \textbf{Majority of $25$-NN} and \textbf{Majority of $50$-NN} indicates for each point in ground truth, we find $25$ or $50$ spatially nearest voxels and treat the majority of their labels as our predictions. After defining our prediction for each point, the rest are totally the same as the evaluation pipeline as proposed in OpenGaussian.
\begin{table}[t]
\centering
 \caption{\textbf{Quantitative evaluation on ScanNet semantic segmentation of $19$ classes.} }
\label{tab:scannet}
\resizebox{0.7\linewidth}{!}{%
\begin{tabular}{l|c|c|c}
\toprule
\textbf{Method} & \textbf{Uses GT Point} & \textbf{mIoU} & \textbf{mAcc}\\
\midrule
LEGaussian~\cite{shi2024legaussian} & \checkmark & 3.8 & 10.9\\
LangSplat~\cite{langsplat} & \checkmark & 3.8 & 9.1 \\
OpenGaussian~\cite{wu2024opengaussian} & \checkmark & 24.7 & 41.5 
\\

\midrule
\textbf{Ours (Nearest)} & - & 30.0 & 41.1\\
\textbf{Ours (Majority of $25$-NN)} & - &31.3 & 42.1\\
\textbf{Ours (Majority of $50$-NN)} & - &\textbf{31.6} & \textbf{42.3}\\
\bottomrule
\end{tabular}}
\end{table}
\paragraph{Results.}
The results are shown in Table~\ref{tab:scannet}. We can see that even without using the ground truth points as prior, our \nickname still outperforms all baselines in terms of mIoU, and is comparable in mAcc. This shows the potential of \nickname on diverse tasks instead of just OVS and RES for being a training-free approach.  

\section{Ablation study.} \label{supp:abla}
Being a training-free approach, it is essential to investigate how different prior models affect the performance of our \nickname. Therefore, we conduct ablation studies in three main prior models: the segmentation model for grouping, the captioning model for generating raw captions, the MLLM for canonical captioning, query refinement, and target retrieval.

\paragraph{Different segmentation model.}
\begin{table}[t]
\centering
 \caption{\textbf{Ablation studies on different segmentation models for RES task on Ref-LeRF~\cite{refersplat} subset.} }
\label{tab:ablation_seg}
\resizebox{0.5\linewidth}{!}{%
\begin{tabular}{l|c|c}
\toprule
\textbf{Method} & \textbf{Segmentation Model} & \textbf{mIoU}\\
\midrule
A & SAM & 30.5\\
B & SAM2 & 42.4 \\
\bottomrule
\end{tabular}}
\end{table}
Table~\ref{tab:ablation_seg} shows our \nickname using different versions of SAM~\cite{sam, ravi2024sam2} as a segmentation prior model for our grouping stage for RES task. In our experiments, we observe that SAM tends to segment small fragments that are over-detailed, and therefore, the grouping results are slightly noisier than our original version using SAM2. As a result, we can see that there is about $10$\% performance drop on the RES task. 

\paragraph{Different captioning model.}
\begin{table}[t]
\centering
 \caption{\textbf{Ablation studies on different models for captioning of RES task on Ref-LeRF~\cite{refersplat} subset.} }
\label{tab:ablation_caption}
\resizebox{0.5\linewidth}{!}{%
\begin{tabular}{l|c|c}
\toprule
\textbf{Method} & \textbf{Captioning Model} & \textbf{mIoU}\\
\midrule
A & Osprey (Yuan, et al, 2024) & 29.3\\
B & Qwen3-VL-8B-Instruct &  33.3\\
C & DAM~\cite{lian2025dam} &  42.4\\
\bottomrule
\end{tabular}}
\end{table}
Table~\ref{tab:ablation_caption} shows our \nickname using different captioning model to obtain the original caption for each group on the RES task. For the setting using the Osprey (Yuan, et al, 2024) captioning model, we caption each frame and then ask Qwen3-VL-8B-Instruct model to summarize them into one sentence since Osprey is not suitable for taking video as input. As for the setting using Qwen3-VL-8B-Instruct as captioning model, we direct bypass the DAM captioning stage and ask Qwen3-VL-8B-Instruct to generate caption purely from the visual input (with darkened background and red dot as visual prompt) since it is not trained for taking separated video-mask pair as input. From Table~\ref{tab:ablation_caption} we can see that although Qwen3-VL-8B-Instruct is not trained specially for captioning masked region captioning task, it can still produce reasonable results that are feasible for the RES task. As for Osprey, although it is a captioning specialized model, the per-frame prediction property lead to inconsistent caption for the same group from different views, confusing the Qwen3-VL-8B-Instruct model for summarization and hindering the performance of the RES task. However, we note that using either captioning model achieves better mIoU than ReferSplat~\cite{refersplat} ($29.2$\% for the original reported number and $24.5$\% for our reproduced results) on the RES task, showing the robustness of our designed pipeline.  
\paragraph{Different MLLM.}
\begin{table}[t]
\centering
 \caption{\textbf{Ablation studies on different MLLMs for RES task on Ref-LeRF~\cite{refersplat} subset.} }
\label{tab:ablation_mllm}
\resizebox{0.5\linewidth}{!}{%
\begin{tabular}{l|c|c}
\toprule
\textbf{Method} & \textbf{MLLM} & \textbf{mIoU}\\
\midrule
A & Qwen2.5-VL-7B-Instruct & 23.4\\
B & Qwen3-VL-2B-Instruct &  10.0\\
C & Qwen3-VL-4B-Instruct &  35.6\\
D &Qwen3-VL-8B-Instruct& 42.4 \\
\bottomrule
\end{tabular}}
\end{table}
Table~\ref{tab:ablation_mllm} shows the results of our \nickname using different MLLMs during the canonical scene map construction and the inference stage with the totally same system prompts and user prompts. Since we do not specially design different prompt for each model, we observe that the Qwen3-VL-2B-Instruct
is incapable of canonicalize the captions generated from DAM. Instead, it tends to repeat some of the words in the original caption as the refined caption. As a result, the incorrect refined captions are hard for the Qwen3-VL-2B-Instruct model to locate the correct target object in the inference stage, leading to a catastrophic $9.98$\% mIoU. In contrast, the newest and largest model for this ablation, Qwen3-VL-8B-Instruct is obviously performing best.
\section{Qualitative results} \label{supp:qual}

\paragraph{Referring Segmentation.}
\begin{figure*}[!t]
    \centering
    \includegraphics[width=1.0\textwidth,]{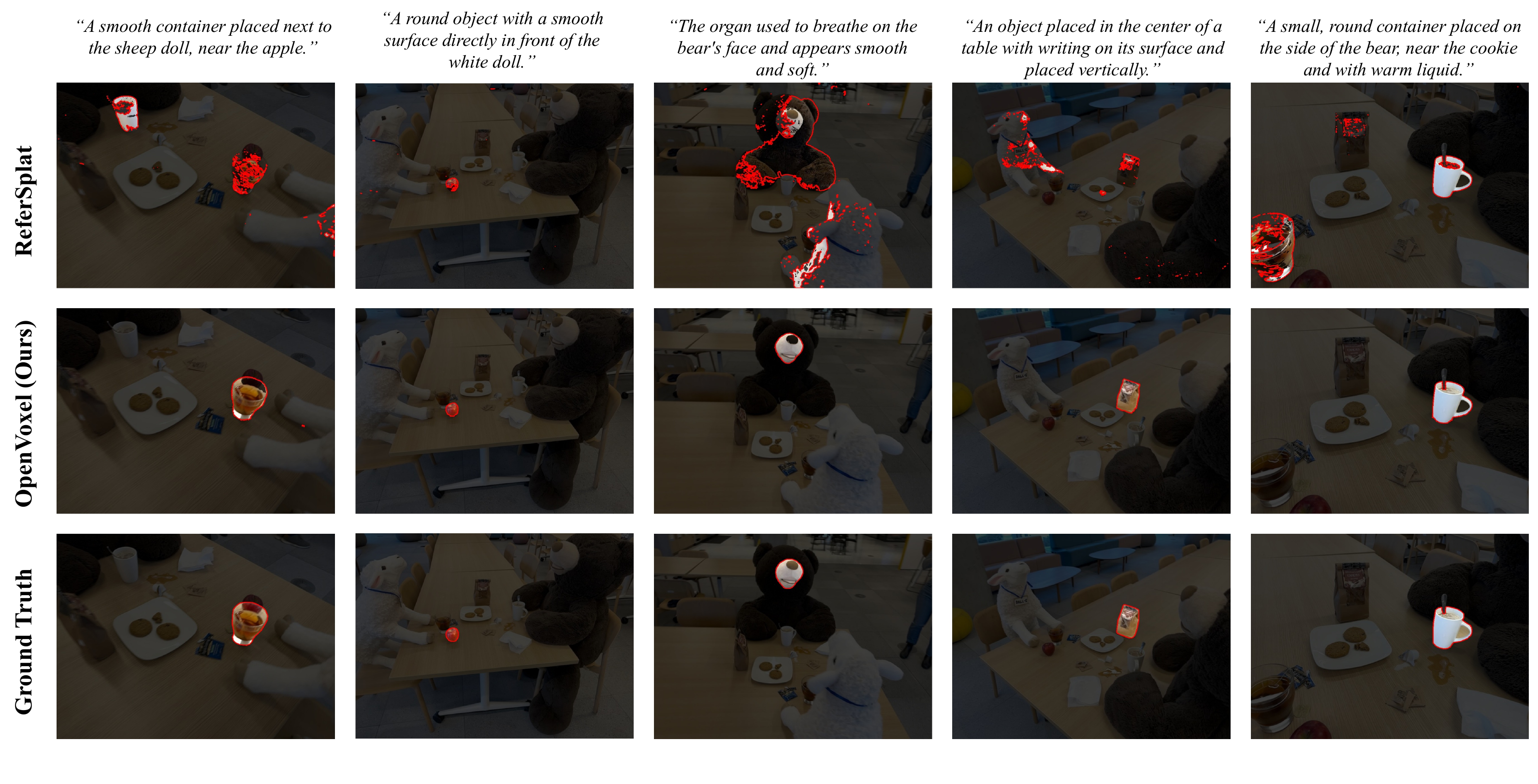}
    \caption{\textbf{Qualitative results of RES task on the \textit{Teatime} scene.} }
    \label{fig:teatime_res}
\end{figure*}

\begin{figure*}[!t]
    \centering
    \includegraphics[width=1.0\textwidth,]{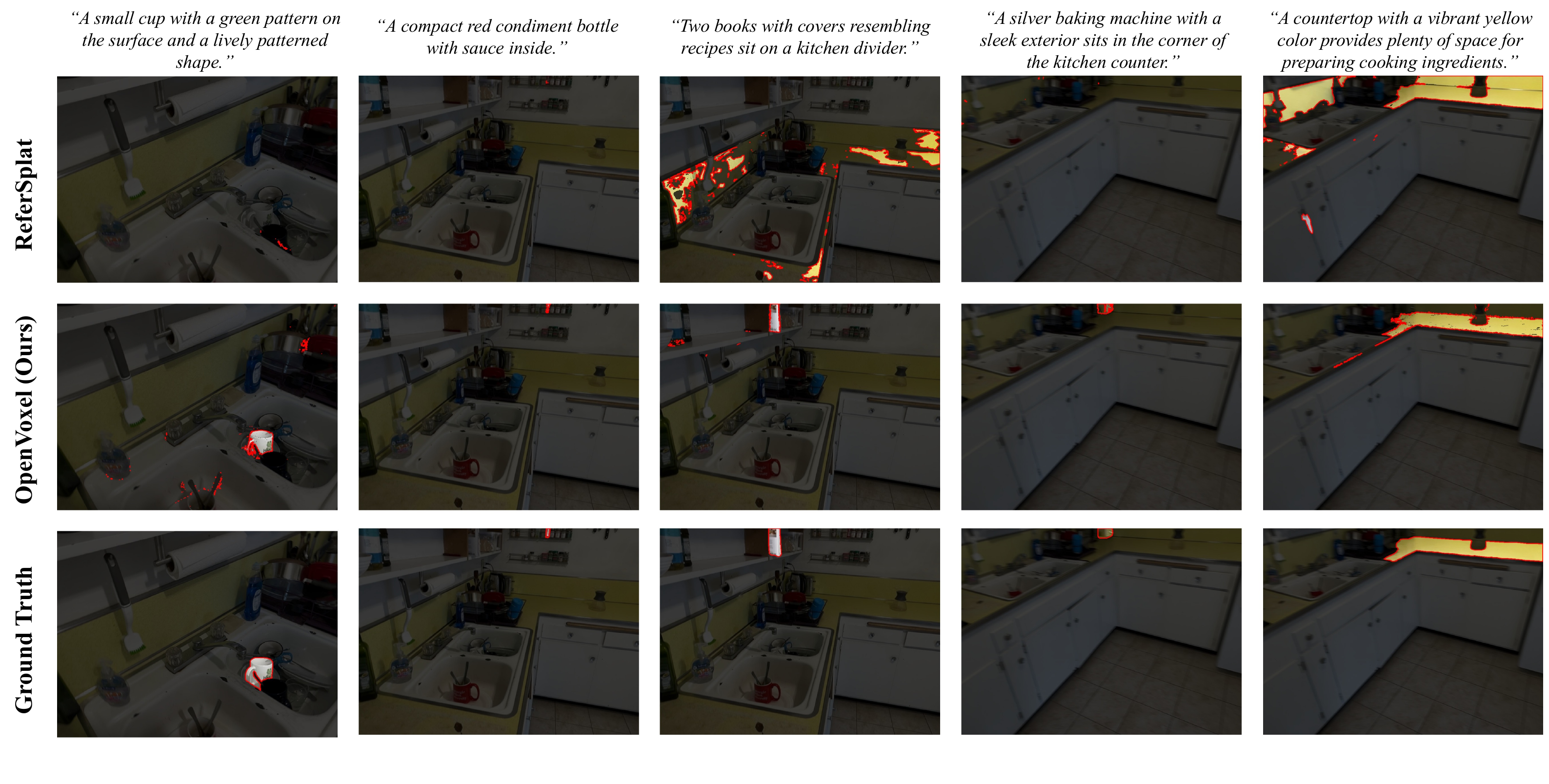}
    \caption{\textbf{Qualitative results of RES task on the \textit{Kitchen} scene.} }
    \label{fig:kitchen_res}
\end{figure*}

We provide the qualitative results of the other two scenes (i.e., teatime and kitchen) in Ref-LeRF~\cite{refersplat} subset for RES task in Fig.~\ref{fig:teatime_res} and Fig.~\ref{fig:kitchen_res}. Similar to our observation in Sect.~\ref{subsec:qualitative}, for the first column (i.e., ``A smooth container placed next to the sheep doll, near the apple'' as query) in Fig.~\ref{fig:teatime_res}, ReferSplat~\cite{refersplat} tends to capture only part of the query (i.e., ``container'') and hence segmenting both the coffe mug and the glass of tea, neglecting other spatial clues in the query. Similarly, for the last column in Fig.~\ref{fig:kitchen_res} with ``A countertop with a vibrant yellow
color provides plenty of space for
preparing cooking ingredients.'' as query, ReferSplat only capture the color information of ``vibrant yellow'' and segment both the counter top and the wall. In contrast, our \nickname successfully locates the ideal target in both cases.

\begin{figure*}[!t]
    \centering
    \includegraphics[width=1.0\textwidth,]{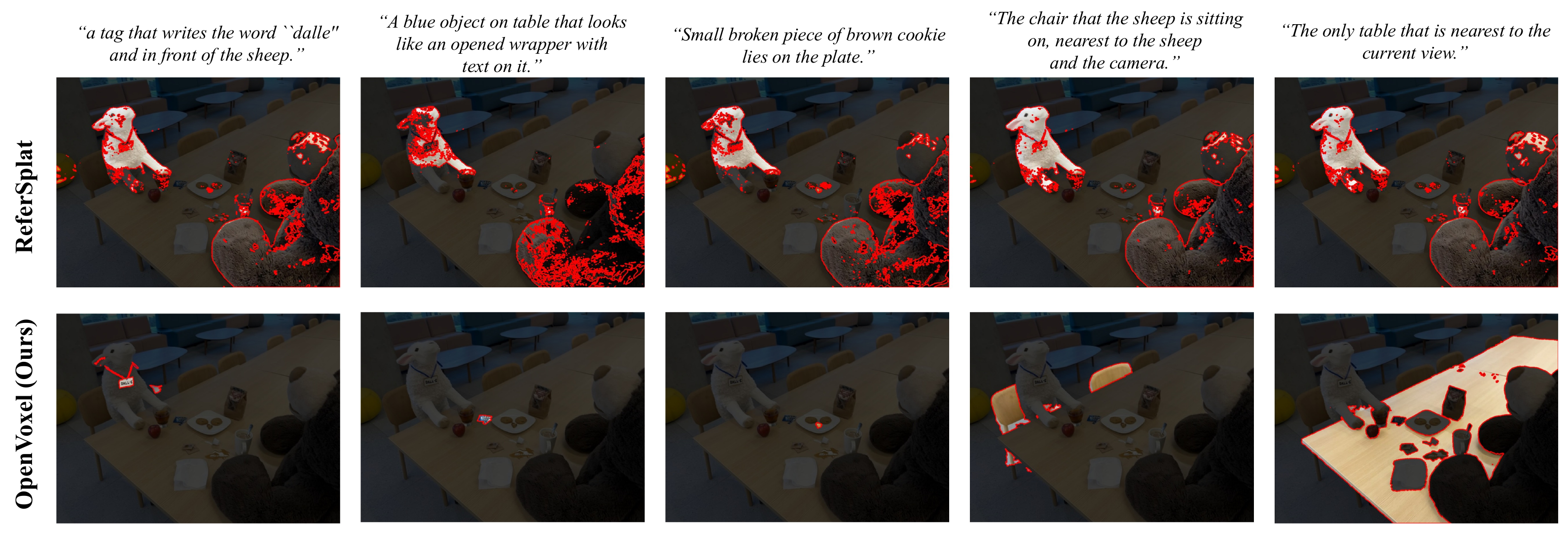}
    \caption{\textbf{Qualitative results of RES task on the \textit{Teatime} scene with other queries.} Note that these queries are created additionally and all of them are not appeared in the original annotations from Ref-LeRF subset (so there are no ground truth mask for them). We can see that ReferSplat~\cite{refersplat} struggles to recognize unseen target objects even in the same scene it is optimized, showing that it tends to overfit on annotated objects from the dataset.}
    \label{fig:teatime_res_ours}
\end{figure*}

We additionally showcase the qualitative results of RES on the ``\textit{Teatime}'' scene using our created natural language that are not included in the Ref-LeRF subset as query to demonstrate the capability of our \nickname compared with ReferSplat~\cite{refersplat}. We can see that since ReferSplat requires training on all objects using human annotated sentences, it is not able to locate objects that are not annotated during their training. In contrast, our training-free approach is not depending on the training annotations at all and is able to locate the correct targets. We note that for the last two columns where the input query contains view dependent descriptions, although our \nickname retrieve two targets instead of the only one matched, the results are still including the correct target, showing the potential capability of solving view-specific tasks.
\section{Discussions and Limitations.}
\label{sec:limiation}
We now discuss the potential limitation of our \nickname. As a training-free approach, the grouping process of \nickname is relatively sensitive to parameters comparing to the end-to-end generalizable ones~\cite{chou2024gsnerf}. As described in Sect.~\ref{supp:details}, the sampling rate of frames and merging frequency are customized for each scene. And since how well-separated for different instances largely affects the performance for both OVS and RES (semantic segmentation is less affected), the SAM2 parameters are needed to be adjust carefully. However, we note that other optimization-based grouping methods~\cite{ye2023gaussiangrouping, zhu2025objectgs} also share similar limitation, as their results heavily rely on the quality of view-consistent video segmentation models such as DEVA~\cite{cheng2023deva}. Fortunately, our heuristic of sampling one frame per 3-5 frames and conduct merging once per 3 steps generally work well. In case the result is unsatisfactory, our work is totally training-free so it would be easy to adjust the parameters and re-run the grouping process with a tolerable time (about 1 minute per-scene, taking Ref-LeRF subset as example.)

Also, since we conduct the instance-level grouping process before captioning/retrieval, if the user gives a query to indicate some part of a larger object (e.g., flash light of the camera in Fig.~\ref{fig:ref_figurines}), our \nickname would still segment the whole object (i.e., the whole camera) since the small parts of the same object is bundled together. One possible solution to solve this issue is to curate the groups as small as possible while still keeping them semantically reasonable (requires 2D segmentation maps that includes those small part). Additionally, instead of just build the Scene Map $S$ by storing center locations of each group, construct a complex scene graph for all the groups to indicate the spatial relations (e.g., on top of, between) or ownership (e.g., belongs to, part of) explicitly. We truly believes that this would help improving the performance and robustness of our \nickname and leave it as a possible future direction.

Another limitation of our \nickname lies in the usage of MLLM models. As shown in the Sect.~\ref{supp:details}, we turn off the chain-of-thought process of the MLLM model for fast inference (less than one second per query). However, by doing so the capability of reasoning for the MLLM is also limited, and hence both the canonicalized captions and the retrieving process are having room to be improved. Also, as shown in Table~\ref{tab:ablation_mllm}, the results of using different open-source MLLM differs. We believe that if better open-source MLLM appear with faster thinking/reasoning ability, our \nickname can be benefited from them.


\clearpage
\setcitestyle{numbers}
\bibliographystyle{plainnat}
\bibliography{main}

\begin{thebibliography}{77}
\providecommand{\natexlab}[1]{#1}
\providecommand{\url}[1]{\texttt{#1}}
\expandafter\ifx\csname urlstyle\endcsname\relax
  \providecommand{\doi}[1]{doi: #1}\else
  \providecommand{\doi}{doi: \begingroup \urlstyle{rm}\Url}\fi

\bibitem[Bai et~al.(2023)Bai, Bai, Yang, Wang, Tan, Wang, Lin, Zhou, and Zhou]{Qwen-VL}
Jinze Bai, Shuai Bai, Shusheng Yang, Shijie Wang, Sinan Tan, Peng Wang, Junyang Lin, Chang Zhou, and Jingren Zhou.
\newblock Qwen-vl: A versatile vision-language model for understanding, localization, text reading, and beyond.
\newblock \emph{arXiv preprint arXiv:2308.12966}, 2023.

\bibitem[Bai et~al.(2025)Bai, Chen, Liu, Wang, Ge, Song, Dang, Wang, Wang, Tang, Zhong, Zhu, Yang, Li, Wan, Wang, Ding, Fu, Xu, Ye, Zhang, Xie, Cheng, Zhang, Yang, Xu, and Lin]{Qwen2.5-VL}
Shuai Bai, Keqin Chen, Xuejing Liu, Jialin Wang, Wenbin Ge, Sibo Song, Kai Dang, Peng Wang, Shijie Wang, Jun Tang, Humen Zhong, Yuanzhi Zhu, Mingkun Yang, Zhaohai Li, Jianqiang Wan, Pengfei Wang, Wei Ding, Zheren Fu, Yiheng Xu, Jiabo Ye, Xi~Zhang, Tianbao Xie, Zesen Cheng, Hang Zhang, Zhibo Yang, Haiyang Xu, and Junyang Lin.
\newblock Qwen2.5-vl technical report.
\newblock \emph{arXiv preprint arXiv:2502.13923}, 2025.

\bibitem[Barron et~al.(2021)Barron, Mildenhall, Tancik, Hedman, Martin-Brualla, and Srinivasan]{barron2021mip}
Jonathan~T Barron, Ben Mildenhall, Matthew Tancik, Peter Hedman, Ricardo Martin-Brualla, and Pratul~P Srinivasan.
\newblock Mip-nerf: A multiscale representation for anti-aliasing neural radiance fields.
\newblock In \emph{Proceedings of the IEEE International Conference on Computer Vision (ICCV)}, 2021.

\bibitem[Cen et~al.(2023)Cen, Zhou, Fang, Shen, Xie, Jiang, Zhang, Tian, et~al.]{cen2023sa3d}
Jiazhong Cen, Zanwei Zhou, Jiemin Fang, Wei Shen, Lingxi Xie, Dongsheng Jiang, Xiaopeng Zhang, Qi~Tian, et~al.
\newblock Segment anything in 3d with nerfs.
\newblock \emph{Advances in Neural Information Processing Systems (NeurIPS)}, 2023.

\bibitem[Chen et~al.(2022)Chen, Xu, Geiger, Yu, and Su]{chen2022tensorf}
Anpei Chen, Zexiang Xu, Andreas Geiger, Jingyi Yu, and Hao Su.
\newblock Tensorf: Tensorial radiance fields.
\newblock In \emph{Proceedings of the European Conference on Computer Vision (ECCV)}, 2022.

\bibitem[Chen and Wang(2024)]{chen2024surveygs}
Guikun Chen and Wenguan Wang.
\newblock A survey on 3d gaussian splatting.
\newblock \emph{arXiv preprint arXiv:2401.03890}, 2024.

\bibitem[Chen et~al.(2024{\natexlab{a}})Chen, Loy, and Pan]{chen2024mvip}
Honghua Chen, Chen~Change Loy, and Xingang Pan.
\newblock Mvip-nerf: Multi-view 3d inpainting on nerf scenes via diffusion prior.
\newblock In \emph{Proceedings of the IEEE Conference on Computer Vision and Pattern Recognition (CVPR)}, 2024{\natexlab{a}}.

\bibitem[Chen et~al.(2024{\natexlab{b}})Chen, Chen, Zhang, Wang, Yang, Wang, Cai, Yang, Liu, and Lin]{chen2024gaussianeditor}
Yiwen Chen, Zilong Chen, Chi Zhang, Feng Wang, Xiaofeng Yang, Yikai Wang, Zhongang Cai, Lei Yang, Huaping Liu, and Guosheng Lin.
\newblock Gaussianeditor: Swift and controllable 3d editing with gaussian splatting.
\newblock In \emph{Proceedings of the IEEE Conference on Computer Vision and Pattern Recognition (CVPR)}, 2024{\natexlab{b}}.

\bibitem[Cheng et~al.(2023)Cheng, Oh, Price, Schwing, and Lee]{cheng2023deva}
Ho~Kei Cheng, Seoung~Wug Oh, Brian Price, Alexander Schwing, and Joon-Young Lee.
\newblock Tracking anything with decoupled video segmentation.
\newblock In \emph{Proceedings of the IEEE International Conference on Computer Vision (ICCV)}, 2023.

\bibitem[Chou et~al.(2024)Chou, Huang, Liu, Wang, et~al.]{chou2024gsnerf}
Zi-Ting Chou, Sheng-Yu Huang, I~Liu, Yu-Chiang~Frank Wang, et~al.
\newblock Gsnerf: Generalizable semantic neural radiance fields with enhanced 3d scene understanding.
\newblock In \emph{Proceedings of the IEEE Conference on Computer Vision and Pattern Recognition (CVPR)}, 2024.

\bibitem[Dai et~al.(2017)Dai, Chang, Savva, Halber, Funkhouser, and Nie{\ss}ner]{dai2017scannet}
Angela Dai, Angel~X Chang, Manolis Savva, Maciej Halber, Thomas Funkhouser, and Matthias Nie{\ss}ner.
\newblock Scannet: Richly-annotated 3d reconstructions of indoor scenes.
\newblock In \emph{Proceedings of the IEEE Conference on Computer Vision and Pattern Recognition (CVPR)}, 2017.

\bibitem[Fridovich-Keil et~al.(2022)Fridovich-Keil, Yu, Tancik, Chen, Recht, and Kanazawa]{fridovich2022plenoxels}
Sara Fridovich-Keil, Alex Yu, Matthew Tancik, Qinhong Chen, Benjamin Recht, and Angjoo Kanazawa.
\newblock Plenoxels: Radiance fields without neural networks.
\newblock In \emph{Proceedings of the IEEE Conference on Computer Vision and Pattern Recognition (CVPR)}, 2022.

\bibitem[Gao et~al.(2024)Gao, Gu, Lin, Zhu, Cao, Zhang, and Yao]{gao2023relightable}
Jian Gao, Chun Gu, Youtian Lin, Hao Zhu, Xun Cao, Li~Zhang, and Yao Yao.
\newblock Relightable 3d gaussian: Real-time point cloud relighting with brdf decomposition and ray tracing.
\newblock \emph{Proceedings of the European Conference on Computer Vision (ECCV)}, 2024.

\bibitem[Haque et~al.(2023)Haque, Tancik, Efros, Holynski, and Kanazawa]{haque2023instruct}
Ayaan Haque, Matthew Tancik, Alexei~A Efros, Aleksander Holynski, and Angjoo Kanazawa.
\newblock Instruct-nerf2nerf: Editing 3d scenes with instructions.
\newblock In \emph{Proceedings of the IEEE International Conference on Computer Vision (ICCV)}, 2023.

\bibitem[He et~al.(2025{\natexlab{a}})He, Peng, Jiang, Hu, and Zhang]{he2025pointseg}
Qingdong He, Jinlong Peng, Zhengkai Jiang, Xiaobin Hu, and Jiangning Zhang.
\newblock Pointseg: A training-free paradigm for 3d scene segmentation via foundation models.
\newblock In \emph{Proceedings of the IEEE International Conference on Computer Vision Workshops (ICCV Workshops)}, 2025{\natexlab{a}}.

\bibitem[He et~al.(2025{\natexlab{b}})He, Jie, Wang, Zhou, Hu, Li, and Ding]{refersplat}
Shuting He, Guangquan Jie, Changshuo Wang, Yun Zhou, Shuming Hu, Guanbin Li, and Henghui Ding.
\newblock Refersplat: Referring segmentation in 3d gaussian splatting.
\newblock \emph{Proceedings of the International Conference on Machine Learning (ICML)}, 2025{\natexlab{b}}.

\bibitem[Hu et~al.(2024)Hu, Fu, Guo, Peng, Chu, Liu, Liu, and Gong]{hu2024innout}
Dongting Hu, Huan Fu, Jiaxian Guo, Liuhua Peng, Tingjin Chu, Feng Liu, Tongliang Liu, and Mingming Gong.
\newblock In-n-out: Lifting 2d diffusion prior for 3d object removal via tuning-free latents alignment.
\newblock \emph{Advances in Neural Information Processing Systems}, 37:\penalty0 45737--45766, 2024.

\bibitem[Huang et~al.(2025{\natexlab{a}})Huang, Chou, and Wang]{huang20253dgic}
Sheng-Yu Huang, Zi-Ting Chou, and Yu-Chiang~Frank Wang.
\newblock 3d gaussian inpainting with depth-guided cross-view consistency.
\newblock In \emph{Proceedings of the IEEE Conference on Computer Vision and Pattern Recognition (CVPR)}, 2025{\natexlab{a}}.

\bibitem[Huang et~al.(2025{\natexlab{b}})Huang, Chen, Hu, Huang, Gong, and Liu]{huang2025openinsgaussian}
Tianyu Huang, Runnan Chen, Dongting Hu, Fengming Huang, Mingming Gong, and Tongliang Liu.
\newblock Openinsgaussian: Open-vocabulary instance gaussian segmentation with context-aware cross-view fusion.
\newblock In \emph{Proceedings of the IEEE International Conference on Computer Vision Workshops (ICCV Workshops)}, 2025{\natexlab{b}}.

\bibitem[Jang and Kim(2025)]{jang2025identityawaregs}
SungMin Jang and Wonjun Kim.
\newblock Identity-aware language gaussian splatting for open-vocabulary 3d semantic segmentation.
\newblock In \emph{Proceedings of the IEEE International Conference on Computer Vision (ICCV)}, 2025.

\bibitem[Ji et~al.(2025)Ji, Zhu, Tang, Liu, Zhang, Tan, and Xie]{ji2025fastlgs}
Yuzhou Ji, He~Zhu, Junshu Tang, Wuyi Liu, Zhizhong Zhang, Xin Tan, and Yuan Xie.
\newblock Fastlgs: Speeding up language embedded gaussians with feature grid mapping.
\newblock In \emph{Proceedings of the AAAI Conference on Artificial Intelligence (AAAI)}, 2025.

\bibitem[Johari et~al.(2022)Johari, Lepoittevin, and Fleuret]{johari2022geonerf}
Mohammad~Mahdi Johari, Yann Lepoittevin, and Fran{\c{c}}ois Fleuret.
\newblock Geonerf: Generalizing nerf with geometry priors.
\newblock In \emph{Proceedings of the IEEE Conference on Computer Vision and Pattern Recognition (CVPR)}, 2022.

\bibitem[Jun-Seong et~al.(2025)Jun-Seong, Kim, Yu-Ji, Wang, Choe, and Oh]{jun2025drsplat}
Kim Jun-Seong, GeonU Kim, Kim Yu-Ji, Yu-Chiang~Frank Wang, Jaesung Choe, and Tae-Hyun Oh.
\newblock Dr. splat: Directly referring 3d gaussian splatting via direct language embedding registration.
\newblock In \emph{Proceedings of the IEEE Conference on Computer Vision and Pattern Recognition (CVPR)}, 2025.

\bibitem[Kamath et~al.(2024)Kamath, Hsieh, Chang, and Krishna]{kamath2024hard}
Amita Kamath, Cheng-Yu Hsieh, Kai-Wei Chang, and Ranjay Krishna.
\newblock The hard positive truth about vision-language compositionality.
\newblock In \emph{Proceedings of the European Conference on Computer Vision (ECCV)}. Springer, 2024.

\bibitem[Kerbl et~al.(2023)Kerbl, Kopanas, Leimk{\"u}hler, and Drettakis]{kerbl202333dgs}
Bernhard Kerbl, Georgios Kopanas, Thomas Leimk{\"u}hler, and George Drettakis.
\newblock 3d gaussian splatting for real-time radiance field rendering.
\newblock \emph{ACM Transactions on Graphics (TOG)}, 2023.

\bibitem[Kerr et~al.(2023)Kerr, Kim, Goldberg, Kanazawa, and Tancik]{lerf}
Justin Kerr, Chung~Min Kim, Ken Goldberg, Angjoo Kanazawa, and Matthew Tancik.
\newblock Lerf: Language embedded radiance fields.
\newblock In \emph{Proceedings of the IEEE International Conference on Computer Vision (ICCV)}, 2023.

\bibitem[Kirillov et~al.(2023)Kirillov, Mintun, Ravi, Mao, Rolland, Gustafson, Xiao, Whitehead, Berg, Lo, et~al.]{sam}
Alexander Kirillov, Eric Mintun, Nikhila Ravi, Hanzi Mao, Chloe Rolland, Laura Gustafson, Tete Xiao, Spencer Whitehead, Alexander~C Berg, Wan-Yen Lo, et~al.
\newblock Segment anything.
\newblock In \emph{Proceedings of the IEEE International Conference on Computer Vision (ICCV)}, 2023.

\bibitem[Kundu et~al.(2022)Kundu, Genova, Yin, Fathi, Pantofaru, Guibas, Tagliasacchi, Dellaert, and Funkhouser]{kundu2022panoptic}
Abhijit Kundu, Kyle Genova, Xiaoqi Yin, Alireza Fathi, Caroline Pantofaru, Leonidas~J Guibas, Andrea Tagliasacchi, Frank Dellaert, and Thomas Funkhouser.
\newblock Panoptic neural fields: A semantic object-aware neural scene representation.
\newblock In \emph{Proceedings of the IEEE Conference on Computer Vision and Pattern Recognition (CVPR)}, 2022.

\bibitem[Lee et~al.(2025)Lee, Park, Choe, Wang, Kautz, Cho, and Choy]{lee2025mosaic3d}
Junha Lee, Chunghyun Park, Jaesung Choe, Yu-Chiang~Frank Wang, Jan Kautz, Minsu Cho, and Chris Choy.
\newblock Mosaic3d: Foundation dataset and model for open-vocabulary 3d segmentation.
\newblock In \emph{Proceedings of the IEEE Conference on Computer Vision and Pattern Recognition (CVPR)}, 2025.

\bibitem[Li et~al.(2025)Li, Zhang, Zhang, Bai, Zheng, Yu, and Gu]{li2025geosvr}
Jiahe Li, Jiawei Zhang, Youmin Zhang, Xiao Bai, Jin Zheng, Xiaohan Yu, and Lin Gu.
\newblock Geosvr: Taming sparse voxels for geometrically accurate surface reconstruction.
\newblock \emph{arXiv preprint arXiv:2509.18090}, 2025.

\bibitem[Lian et~al.(2025)Lian, Ding, Ge, Liu, Mao, Li, Pavone, Liu, Darrell, Yala, et~al.]{lian2025dam}
Long Lian, Yifan Ding, Yunhao Ge, Sifei Liu, Hanzi Mao, Boyi Li, Marco Pavone, Ming-Yu Liu, Trevor Darrell, Adam Yala, et~al.
\newblock Describe anything: Detailed localized image and video captioning.
\newblock \emph{arXiv preprint arXiv:2504.16072}, 2025.

\bibitem[Liao et~al.(2025)Liao, Li, Zheng, Liu, Gao, and Wang]{3dgsgrounding}
Liwei Liao, Xufeng Li, Xiaoyun Zheng, Boning Liu, Feng Gao, and Ronggang Wang.
\newblock Zero-shot visual grounding in 3d gaussians via view retrieval.
\newblock \emph{arXiv preprint arXiv:2509.15871}, 2025.

\bibitem[Lin et~al.(2024)Lin, Kim, Huang, Li, Ma, Kopf, Yang, and Tseng]{lin2024maldnerf}
Chieh~Hubert Lin, Changil Kim, Jia-Bin Huang, Qinbo Li, Chih-Yao Ma, Johannes Kopf, Ming-Hsuan Yang, and Hung-Yu Tseng.
\newblock Taming latent diffusion model for neural radiance field inpainting.
\newblock \emph{Proceedings of the European Conference on Computer Vision (ECCV)}, 2024.

\bibitem[Liu et~al.(2023)Liu, Zhan, Zhang, Xu, Yu, El~Saddik, Theobalt, Xing, and Lu]{3dovs}
Kunhao Liu, Fangneng Zhan, Jiahui Zhang, Muyu Xu, Yingchen Yu, Abdulmotaleb El~Saddik, Christian Theobalt, Eric Xing, and Shijian Lu.
\newblock Weakly supervised 3d open-vocabulary segmentation.
\newblock \emph{Advances in Neural Information Processing Systems (NeurIPS)}, 2023.

\bibitem[Liu et~al.(2022)Liu, Peng, Liu, Wang, Wang, Theobalt, Zhou, and Wang]{liu2022neural}
Yuan Liu, Sida Peng, Lingjie Liu, Qianqian Wang, Peng Wang, Christian Theobalt, Xiaowei Zhou, and Wenping Wang.
\newblock Neural rays for occlusion-aware image-based rendering.
\newblock In \emph{Proceedings of the IEEE Conference on Computer Vision and Pattern Recognition (CVPR)}, 2022.

\bibitem[Liu et~al.(2024)Liu, Ouyang, Wang, Cheng, Xiao, Zhu, Xue, Liu, Shen, and Cao]{liu2024infusion}
Zhiheng Liu, Hao Ouyang, Qiuyu Wang, Ka~Leong Cheng, Jie Xiao, Kai Zhu, Nan Xue, Yu~Liu, Yujun Shen, and Yang Cao.
\newblock Infusion: Inpainting 3d gaussians via learning depth completion from diffusion prior.
\newblock \emph{arXiv preprint arXiv:2404.11613}, 2024.

\bibitem[Lyu et~al.(2024)Lyu, Li, Kundu, Tsai, and Yang]{lyu2024gaga}
Weijie Lyu, Xueting Li, Abhijit Kundu, Yi-Hsuan Tsai, and Ming-Hsuan Yang.
\newblock Gaga: Group any gaussians via 3d-aware memory bank.
\newblock \emph{arXiv preprint arXiv:2404.07977}, 2024.

\bibitem[Marrie et~al.(2025)Marrie, M{\'e}n{\'e}gaux, Arbel, Larlus, and Mairal]{marrie2025ludvig}
Juliette Marrie, Romain M{\'e}n{\'e}gaux, Michael Arbel, Diane Larlus, and Julien Mairal.
\newblock Ludvig: Learning-free uplifting of 2d visual features to gaussian splatting scenes.
\newblock In \emph{Proceedings of the IEEE International Conference on Computer Vision (ICCV)}, 2025.

\bibitem[Martin-Brualla et~al.(2021)Martin-Brualla, Radwan, Sajjadi, Barron, Dosovitskiy, and Duckworth]{martin2021nerf}
Ricardo Martin-Brualla, Noha Radwan, Mehdi~SM Sajjadi, Jonathan~T Barron, Alexey Dosovitskiy, and Daniel Duckworth.
\newblock Nerf in the wild: Neural radiance fields for unconstrained photo collections.
\newblock In \emph{Proceedings of the IEEE Conference on Computer Vision and Pattern Recognition (CVPR)}, 2021.

\bibitem[Mildenhall et~al.(2021)Mildenhall, Srinivasan, Tancik, Barron, Ramamoorthi, and Ng]{mildenhall2021nerf}
Ben Mildenhall, Pratul~P Srinivasan, Matthew Tancik, Jonathan~T Barron, Ravi Ramamoorthi, and Ren Ng.
\newblock Nerf: Representing scenes as neural radiance fields for view synthesis.
\newblock \emph{Communications of the ACM}, 65\penalty0 (1):\penalty0 99--106, 2021.

\bibitem[Mirzaei et~al.(2023{\natexlab{a}})Mirzaei, Aumentado-Armstrong, Brubaker, Kelly, Levinshtein, Derpanis, and Gilitschenski]{mirzaei2023referenceinpaint}
Ashkan Mirzaei, Tristan Aumentado-Armstrong, Marcus~A Brubaker, Jonathan Kelly, Alex Levinshtein, Konstantinos~G Derpanis, and Igor Gilitschenski.
\newblock Reference-guided controllable inpainting of neural radiance fields.
\newblock In \emph{Proceedings of the IEEE International Conference on Computer Vision (ICCV)}, 2023{\natexlab{a}}.

\bibitem[Mirzaei et~al.(2023{\natexlab{b}})Mirzaei, Aumentado-Armstrong, Derpanis, Kelly, Brubaker, Gilitschenski, and Levinshtein]{mirzaei2023spin}
Ashkan Mirzaei, Tristan Aumentado-Armstrong, Konstantinos~G Derpanis, Jonathan Kelly, Marcus~A Brubaker, Igor Gilitschenski, and Alex Levinshtein.
\newblock Spin-nerf: Multiview segmentation and perceptual inpainting with neural radiance fields.
\newblock In \emph{Proceedings of the IEEE Conference on Computer Vision and Pattern Recognition (CVPR)}, 2023{\natexlab{b}}.

\bibitem[Mirzaei et~al.(2024)Mirzaei, De~Lutio, Kim, Acuna, Kelly, Fidler, Gilitschenski, and Gojcic]{mirzaei2024reffusion}
Ashkan Mirzaei, Riccardo De~Lutio, Seung~Wook Kim, David Acuna, Jonathan Kelly, Sanja Fidler, Igor Gilitschenski, and Zan Gojcic.
\newblock Reffusion: Reference adapted diffusion models for 3d scene inpainting.
\newblock \emph{arXiv preprint arXiv:2404.10765}, 2024.

\bibitem[M{\"u}ller et~al.(2022)M{\"u}ller, Evans, Schied, and Keller]{muller2022instant}
Thomas M{\"u}ller, Alex Evans, Christoph Schied, and Alexander Keller.
\newblock Instant neural graphics primitives with a multiresolution hash encoding.
\newblock \emph{ACM Transactions on Graphics (ToG)}, 41\penalty0 (4):\penalty0 1--15, 2022.

\bibitem[Neven et~al.(2019)Neven, Brabandere, Proesmans, and Gool]{deepspatialclustering}
Davy Neven, Bert~De Brabandere, Marc Proesmans, and Luc~Van Gool.
\newblock Instance segmentation by jointly optimizing spatial embeddings and clustering bandwidth.
\newblock In \emph{Proceedings of the IEEE Conference on Computer Vision and Pattern Recognition (CVPR)}, 2019.

\bibitem[Paszke et~al.(2019)Paszke, Gross, Massa, Lerer, Bradbury, Chanan, Killeen, Lin, Gimelshein, Antiga, et~al.]{paszke2019pytorch}
Adam Paszke, Sam Gross, Francisco Massa, Adam Lerer, James Bradbury, Gregory Chanan, Trevor Killeen, Zeming Lin, Natalia Gimelshein, Luca Antiga, et~al.
\newblock Pytorch: An imperative style, high-performance deep learning library.
\newblock \emph{Advances in Neural Information Processing Systems (NeurIPS)}, 2019.

\bibitem[Peng et~al.(2025)Peng, Planche, Gao, Zheng, Choudhuri, Chen, Chen, and Wu]{peng20243dvlgs}
Qucheng Peng, Benjamin Planche, Zhongpai Gao, Meng Zheng, Anwesa Choudhuri, Terrence Chen, Chen Chen, and Ziyan Wu.
\newblock 3d vision-language gaussian splatting.
\newblock \emph{Proceedings of the International Conference on Learning Representations (ICLR)}, 2025.

\bibitem[Piekenbrinck et~al.(2025)Piekenbrinck, Schmidt, Hermans, Vaskevicius, Linder, and Leibe]{piekenbrinck2025opensplat3d}
Jens Piekenbrinck, Christian Schmidt, Alexander Hermans, Narunas Vaskevicius, Timm Linder, and Bastian Leibe.
\newblock Opensplat3d: Open-vocabulary 3d instance segmentation using gaussian splatting.
\newblock In \emph{Proceedings of the IEEE Conference on Computer Vision and Pattern Recognition Workshops (CVPR Workshops)}, 2025.

\bibitem[Qi et~al.(2019)Qi, Litany, He, and Guibas]{deephoughvote}
Charles~R. Qi, Or~Litany, Kaiming He, and Leonidas~J. Guibas.
\newblock Deep hough voting for 3d object detection in point clouds.
\newblock In \emph{Proceedings of the IEEE International Conference on Computer Vision (ICCV)}, 2019.

\bibitem[Qin et~al.(2024)Qin, Li, Zhou, Wang, and Pfister]{langsplat}
Minghan Qin, Wanhua Li, Jiawei Zhou, Haoqian Wang, and Hanspeter Pfister.
\newblock Langsplat: 3d language gaussian splatting.
\newblock In \emph{Proceedings of the IEEE Conference on Computer Vision and Pattern Recognition (CVPR)}, 2024.

\bibitem[Qu et~al.(2024)Qu, Dai, Li, Lin, Cao, Zhang, and Ji]{qu2024goi}
Yansong Qu, Shaohui Dai, Xinyang Li, Jianghang Lin, Liujuan Cao, Shengchuan Zhang, and Rongrong Ji.
\newblock Goi: Find 3d gaussians of interest with an optimizable open-vocabulary semantic-space hyperplane.
\newblock In \emph{Proceedings of the 32nd ACM international conference on multimedia}, pages 5328--5337, 2024.

\bibitem[Radford et~al.(2021)Radford, Kim, Hallacy, Ramesh, Goh, Agarwal, Sastry, Askell, Mishkin, Clark, et~al.]{clip}
Alec Radford, Jong~Wook Kim, Chris Hallacy, Aditya Ramesh, Gabriel Goh, Sandhini Agarwal, Girish Sastry, Amanda Askell, Pamela Mishkin, Jack Clark, et~al.
\newblock Learning transferable visual models from natural language supervision.
\newblock In \emph{International conference on machine learning}, pages 8748--8763. PmLR, 2021.

\bibitem[Ravi et~al.(2024)Ravi, Gabeur, Hu, Hu, Ryali, Ma, Khedr, R{\"a}dle, Rolland, Gustafson, et~al.]{ravi2024sam2}
Nikhila Ravi, Valentin Gabeur, Yuan-Ting Hu, Ronghang Hu, Chaitanya Ryali, Tengyu Ma, Haitham Khedr, Roman R{\"a}dle, Chloe Rolland, Laura Gustafson, et~al.
\newblock Sam 2: Segment anything in images and videos.
\newblock \emph{arXiv preprint arXiv:2408.00714}, 2024.

\bibitem[Reiser et~al.(2021)Reiser, Peng, Liao, and Geiger]{reiser2021kilonerf}
Christian Reiser, Songyou Peng, Yiyi Liao, and Andreas Geiger.
\newblock Kilonerf: Speeding up neural radiance fields with thousands of tiny mlps.
\newblock In \emph{Proceedings of the IEEE International Conference on Computer Vision (ICCV)}, 2021.

\bibitem[Ren et~al.(2024)Ren, Liu, Zeng, Lin, Li, Cao, Chen, Huang, Chen, Yan, et~al.]{ren2024groundedsam}
Tianhe Ren, Shilong Liu, Ailing Zeng, Jing Lin, Kunchang Li, He~Cao, Jiayu Chen, Xinyu Huang, Yukang Chen, Feng Yan, et~al.
\newblock Grounded sam: Assembling open-world models for diverse visual tasks.
\newblock \emph{arXiv preprint arXiv:2401.14159}, 2024.

\bibitem[Shi et~al.(2024)Shi, Wang, Duan, and Guan]{shi2024legaussian}
Jin-Chuan Shi, Miao Wang, Hao-Bin Duan, and Shao-Hua Guan.
\newblock Language embedded 3d gaussians for open-vocabulary scene understanding.
\newblock In \emph{Proceedings of the IEEE Conference on Computer Vision and Pattern Recognition (CVPR)}, 2024.

\bibitem[Suhail et~al.(2022)Suhail, Esteves, Sigal, and Makadia]{suhail2022generalizable}
Mohammed Suhail, Carlos Esteves, Leonid Sigal, and Ameesh Makadia.
\newblock Generalizable patch-based neural rendering.
\newblock In \emph{Proceedings of the European Conference on Computer Vision (ECCV)}, 2022.

\bibitem[Sun et~al.(2022)Sun, Sun, and Chen]{sun2022direct}
Cheng Sun, Min Sun, and Hwann-Tzong Chen.
\newblock Direct voxel grid optimization: Super-fast convergence for radiance fields reconstruction.
\newblock In \emph{Proceedings of the IEEE Conference on Computer Vision and Pattern Recognition (CVPR)}, 2022.

\bibitem[Sun et~al.(2025)Sun, Choe, Loop, Ma, and Wang]{svr}
Cheng Sun, Jaesung Choe, Charles Loop, Wei-Chiu Ma, and Yu-Chiang~Frank Wang.
\newblock Sparse voxels rasterization: Real-time high-fidelity radiance field rendering.
\newblock In \emph{Proceedings of the IEEE Conference on Computer Vision and Pattern Recognition (CVPR)}, 2025.

\bibitem[Tian et~al.(2025)Tian, Li, Ma, Yin, Zheng, Huang, Li, Lu, and Jia]{tian2025ccllgs}
Lei Tian, Xiaomin Li, Liqian Ma, Hao Yin, Zirui Zheng, Hefei Huang, Taiqing Li, Huchuan Lu, and Xu~Jia.
\newblock Ccl-lgs: Contrastive codebook learning for 3d language gaussian splatting.
\newblock In \emph{Proceedings of the IEEE International Conference on Computer Vision (ICCV)}, 2025.

\bibitem[Vora et~al.(2021)Vora, Radwan, Greff, Meyer, Genova, Sajjadi, Pot, Tagliasacchi, and Duckworth]{vora2021nesf}
Suhani Vora, Noha Radwan, Klaus Greff, Henning Meyer, Kyle Genova, Mehdi~SM Sajjadi, Etienne Pot, Andrea Tagliasacchi, and Daniel Duckworth.
\newblock Nesf: Neural semantic fields for generalizable semantic segmentation of 3d scenes.
\newblock \emph{arXiv preprint arXiv:2111.13260}, 2021.

\bibitem[Wang et~al.(2024{\natexlab{a}})Wang, Zhang, Abboud, and S{\"u}sstrunk]{wang2024innerf360}
Dongqing Wang, Tong Zhang, Alaa Abboud, and Sabine S{\"u}sstrunk.
\newblock Innerf360: Text-guided 3d-consistent object inpainting on 360-degree neural radiance fields.
\newblock In \emph{Proceedings of the IEEE Conference on Computer Vision and Pattern Recognition (CVPR)}, 2024{\natexlab{a}}.

\bibitem[Wang et~al.(2024{\natexlab{b}})Wang, Bai, Tan, Wang, Fan, Bai, Chen, Liu, Wang, Ge, Fan, Dang, Du, Ren, Men, Liu, Zhou, Zhou, and Lin]{Qwen2-VL}
Peng Wang, Shuai Bai, Sinan Tan, Shijie Wang, Zhihao Fan, Jinze Bai, Keqin Chen, Xuejing Liu, Jialin Wang, Wenbin Ge, Yang Fan, Kai Dang, Mengfei Du, Xuancheng Ren, Rui Men, Dayiheng Liu, Chang Zhou, Jingren Zhou, and Junyang Lin.
\newblock Qwen2-vl: Enhancing vision-language model's perception of the world at any resolution.
\newblock \emph{arXiv preprint arXiv:2409.12191}, 2024{\natexlab{b}}.

\bibitem[Wang et~al.(2024{\natexlab{c}})Wang, Wu, Zhang, and Xu]{wang2024gscream}
Yuxin Wang, Qianyi Wu, Guofeng Zhang, and Dan Xu.
\newblock Gscream: Learning 3d geometry and feature consistent gaussian splatting for object removal.
\newblock \emph{Proceedings of the European Conference on Computer Vision (ECCV)}, 2024{\natexlab{c}}.

\bibitem[Weber et~al.(2024)Weber, Holynski, Jampani, Saxena, Snavely, Kar, and Kanazawa]{weber2024nerfiller}
Ethan Weber, Aleksander Holynski, Varun Jampani, Saurabh Saxena, Noah Snavely, Abhishek Kar, and Angjoo Kanazawa.
\newblock Nerfiller: Completing scenes via generative 3d inpainting.
\newblock In \emph{Proceedings of the IEEE Conference on Computer Vision and Pattern Recognition (CVPR)}, pages 20731--20741, 2024.

\bibitem[Weder et~al.(2023)Weder, Garcia-Hernando, Monszpart, Pollefeys, Brostow, Firman, and Vicente]{weder2023removingnerf}
Silvan Weder, Guillermo Garcia-Hernando, Aron Monszpart, Marc Pollefeys, Gabriel~J Brostow, Michael Firman, and Sara Vicente.
\newblock Removing objects from neural radiance fields.
\newblock In \emph{Proceedings of the IEEE Conference on Computer Vision and Pattern Recognition (CVPR)}, 2023.

\bibitem[Wu et~al.(2025)Wu, Chen, Chen, Lee, Ke, Mu, Huang, Lin, Chen, Lin, et~al.]{wu2025aurafusion360}
Chung-Ho Wu, Yang-Jung Chen, Ying-Huan Chen, Jie-Ying Lee, Bo-Hsu Ke, Chun-Wei~Tuan Mu, Yi-Chuan Huang, Chin-Yang Lin, Min-Hung Chen, Yen-Yu Lin, et~al.
\newblock Aurafusion360: Augmented unseen region alignment for reference-based 360deg unbounded scene inpainting.
\newblock In \emph{Proceedings of the IEEE Conference on Computer Vision and Pattern Recognition (CVPR)}, 2025.

\bibitem[Wu et~al.(2024{\natexlab{a}})Wu, Zhang, Xia, Li, Xia, Chang, Yu, Kim, Rossi, Zhang, et~al.]{wu2024visualprompt}
Junda Wu, Zhehao Zhang, Yu~Xia, Xintong Li, Zhaoyang Xia, Aaron Chang, Tong Yu, Sungchul Kim, Ryan~A Rossi, Ruiyi Zhang, et~al.
\newblock Visual prompting in multimodal large language models: A survey.
\newblock \emph{arXiv preprint arXiv:2409.15310}, 2024{\natexlab{a}}.

\bibitem[Wu et~al.(2024{\natexlab{b}})Wu, Meng, Li, Wu, Shi, Cheng, Zhao, Feng, Ding, Wang, et~al.]{wu2024opengaussian}
Yanmin Wu, Jiarui Meng, Haijie Li, Chenming Wu, Yahao Shi, Xinhua Cheng, Chen Zhao, Haocheng Feng, Errui Ding, Jingdong Wang, et~al.
\newblock Opengaussian: Towards point-level 3d gaussian-based open vocabulary understanding.
\newblock \emph{Advances in Neural Information Processing Systems (NeurIPS)}, 2024{\natexlab{b}}.

\bibitem[Xu et~al.(2025)Xu, Peng, Wang, Blum, Barath, Geiger, and Pollefeys]{xu2025depthsplat}
Haofei Xu, Songyou Peng, Fangjinhua Wang, Hermann Blum, Daniel Barath, Andreas Geiger, and Marc Pollefeys.
\newblock Depthsplat: Connecting gaussian splatting and depth.
\newblock In \emph{Proceedings of the IEEE Conference on Computer Vision and Pattern Recognition (CVPR)}, 2025.

\bibitem[Ye et~al.(2024)Ye, Danelljan, Yu, and Ke]{ye2023gaussiangrouping}
Mingqiao Ye, Martin Danelljan, Fisher Yu, and Lei Ke.
\newblock Gaussian grouping: Segment and edit anything in 3d scenes.
\newblock \emph{Proceedings of the European Conference on Computer Vision (ECCV)}, 2024.

\bibitem[Yin et~al.(2023)Yin, Fu, Yang, and Lin]{yin2023ornerf}
Youtan Yin, Zhoujie Fu, Fan Yang, and Guosheng Lin.
\newblock Or-nerf: Object removing from 3d scenes guided by multiview segmentation with neural radiance fields.
\newblock \emph{arXiv preprint arXiv:2305.10503}, 2023.

\bibitem[Ying et~al.(2024)Ying, Yin, Zhang, Wang, Yu, Huang, and Fang]{ying2024omniseg3d}
Haiyang Ying, Yixuan Yin, Jinzhi Zhang, Fan Wang, Tao Yu, Ruqi Huang, and Lu~Fang.
\newblock Omniseg3d: Omniversal 3d segmentation via hierarchical contrastive learning.
\newblock In \emph{Proceedings of the IEEE Conference on Computer Vision and Pattern Recognition (CVPR)}, 2024.

\bibitem[Yu et~al.(2021)Yu, Li, Tancik, Li, Ng, and Kanazawa]{yu2021plenoctrees}
Alex Yu, Ruilong Li, Matthew Tancik, Hao Li, Ren Ng, and Angjoo Kanazawa.
\newblock Plenoctrees for real-time rendering of neural radiance fields.
\newblock In \emph{Proceedings of the IEEE International Conference on Computer Vision (ICCV)}, 2021.

\bibitem[Yu et~al.(2024)Yu, Chen, Huang, Sattler, and Geiger]{yu2024mipsplat}
Zehao Yu, Anpei Chen, Binbin Huang, Torsten Sattler, and Andreas Geiger.
\newblock Mip-splatting: Alias-free 3d gaussian splatting.
\newblock In \emph{Proceedings of the IEEE Conference on Computer Vision and Pattern Recognition (CVPR)}, 2024.

\bibitem[Zhou et~al.(2024)Zhou, Chang, Jiang, Fan, Zhu, Xu, Chari, You, Wang, and Kadambi]{zhou2024featuregs}
Shijie Zhou, Haoran Chang, Sicheng Jiang, Zhiwen Fan, Zehao Zhu, Dejia Xu, Pradyumna Chari, Suya You, Zhangyang Wang, and Achuta Kadambi.
\newblock Feature 3dgs: Supercharging 3d gaussian splatting to enable distilled feature fields.
\newblock In \emph{Proceedings of the IEEE Conference on Computer Vision and Pattern Recognition (CVPR)}, 2024.

\bibitem[Zhu et~al.(2025)Zhu, Yu, Xu, Jiang, Li, Zhang, Pang, and Dai]{zhu2025objectgs}
Ruijie Zhu, Mulin Yu, Linning Xu, Lihan Jiang, Yixuan Li, Tianzhu Zhang, Jiangmiao Pang, and Bo~Dai.
\newblock Objectgs: Object-aware scene reconstruction and scene understanding via gaussian splatting.
\newblock In \emph{Proceedings of the IEEE International Conference on Computer Vision (ICCV)}, 2025.

\end{thebibliography}

\end{document}